\PassOptionsToPackage{square,comma,numbers,sort&compress}{natbib}
\documentclass{article}

\usepackage{neurips_2019}

\usepackage{xcolor}
\usepackage{graphicx}
\usepackage[utf8]{inputenc} 
\usepackage[T1]{fontenc}    
\usepackage{hyperref}       
\usepackage{url}            
\usepackage{booktabs}       
\usepackage{amsfonts}       
\usepackage{amsmath}
\usepackage{amssymb}
\usepackage{authblk}
\usepackage{caption}
\usepackage{subcaption}

\usepackage{bm}
\DeclareMathOperator*{\argmax}{arg\,max}
\DeclareMathOperator*{\argmin}{arg\,min}

\usepackage{nicefrac}       
\usepackage{microtype}      

\usepackage{natbib}

\title{Robust Sparse Regularization: Simultaneously Optimizing Neural Network Robustness and Compactness}

\author[1]{Adnan Siraj Rakin}
\author[1]{Zhezhi He}
\author[1]{Li Yang}
\author[2]{Yanzhi Wang}
\author[3]{Liqiang Wang} 
\author[1]{Deliang Fan}
\affil[1]{Department of Computer Engineering, University of Central Florida \thanks{ Corresponding Author: dfan@ucf.edu}}
\affil[2]{Department of Electrical and Computer Engineering, Northeastern University}
\affil[3]{Department of Computer Science, University of Central Florida}

\begin{document}
\maketitle
\vspace{-1em}
\begin{abstract}

Deep Neural Network (DNN) trained by the gradient descent method is known to be vulnerable to maliciously perturbed adversarial input, aka. adversarial attack. As one of the countermeasures against adversarial attack, increasing the model capacity for DNN robustness enhancement was discussed and reported as an effective approach by many recent works.
In this work, we show that shrinking the model size through proper weight pruning can even be helpful to improve the DNN robustness under adversarial attack. For obtaining a simultaneously robust and compact DNN model, we propose a multi-objective training method called \textit{Robust Sparse Regularization} (RSR), through the fusion of various regularization techniques, including channel-wise noise injection, lasso weight penalty, and adversarial training. 
We conduct extensive experiments across popular ResNet-20, ResNet-18 and VGG-16 DNN architectures to demonstrate the effectiveness of RSR against popular white-box (i.e., PGD and FGSM) and black-box attacks. Thanks to RSR, \textbf{85\%} weight connections of ResNet-18 can be pruned while still achieving \textbf{0.68\%} and \textbf{8.72\%} improvement in clean- and perturbed-data accuracy respectively on CIFAR-10 dataset, in comparison to its PGD adversarial training baseline.

\end{abstract}

\section{Introduction}

Deep Neural Networks (DNNs) have led to tremendous success in various applications, such as image classification \cite{hinton2012neural}, speech recognition \cite{hinton2012deep}, medical applications \cite{hung2017comparing} and etc. Wide deployment of DNNs has raised several major security concerns \cite{goodfellow2014explaining,akhtar2018threat,chen2017show}. For example, in the context of image classification, an adversarial example is a carefully modified image that is visually imperceptible to human eyes, but fools the DNN successfully \cite{goodfellow2014explaining}. 
Recently, there have been a cohort of works toward developing new adversarial attack techniques, which have exposed the underlying vulnerability of DNN \cite{pmlr-v80-athalye18a,madry2018towards}. In order to counter adversarial attacks, several works have proposed different techniques, such as training the network with adversarial samples \cite{madry2018towards,goodfellow2014explaining}, regularization \cite{he2019PNI,lin2018defensive} and other various methods \cite{raghunathan2018certified,samangouei2018defensegan}.

In a separate yet related track, investigation towards generating efficient and compact networks have also been accelerated. Many prior works have been conducted regarding compression techniques including quantization \cite{zhou2016dorefa,courbariaux2015binaryconnect,courbariaux2016binarized,he2019simultaneously} and weight pruning \cite{han2015learning,molchanov2017variational,wen2016learning}. It is shown that many DNNs can function properly (no accuracy loss) even after significantly (>90\%) network pruning \cite{han2015deep,molchanov2017variational}. Such sparse DNN achieves significant speed-up and compression rate which opens the door for DNN in memory and resource constraint applications \cite{han2016mcdnn}. Previously, several works have tried to generate both sparse and robust networks \cite{guo2018sparse,ye2019second} by combining the network pruning (i.e. compactness) technique to defend adversarial examples. However, their efforts either suffer from poor test data accuracy or do not improve the robustness significantly.

\begin{figure}[t]
\centering
\begin{minipage}[c][1\width]{0.3\textwidth} 
			\centering
			\includegraphics[width=0.9\textwidth]{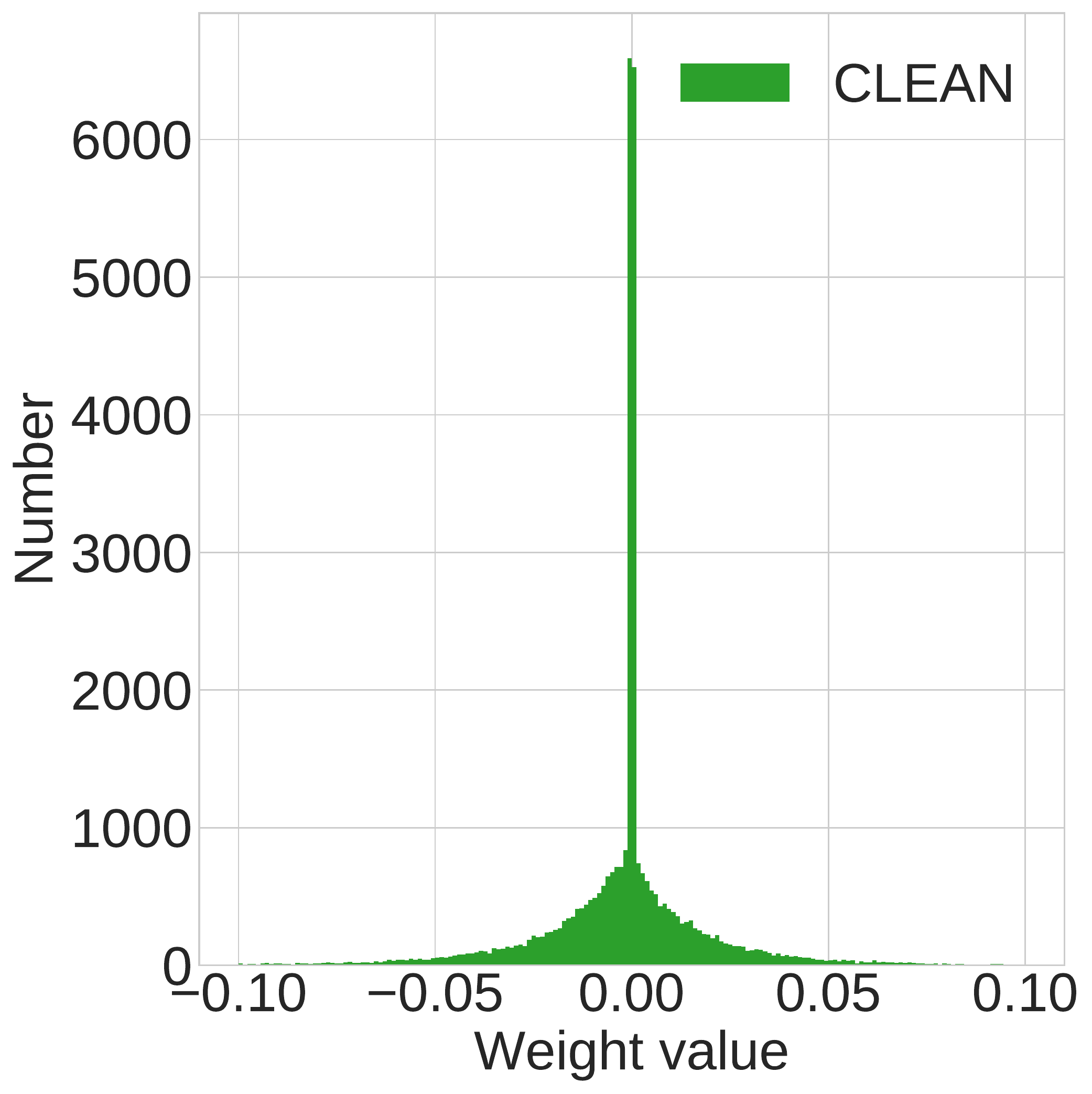} \end{minipage} \quad
\begin{minipage}[c][1\width]{0.3\textwidth} 
			\centering
			\includegraphics[width=0.9\textwidth]{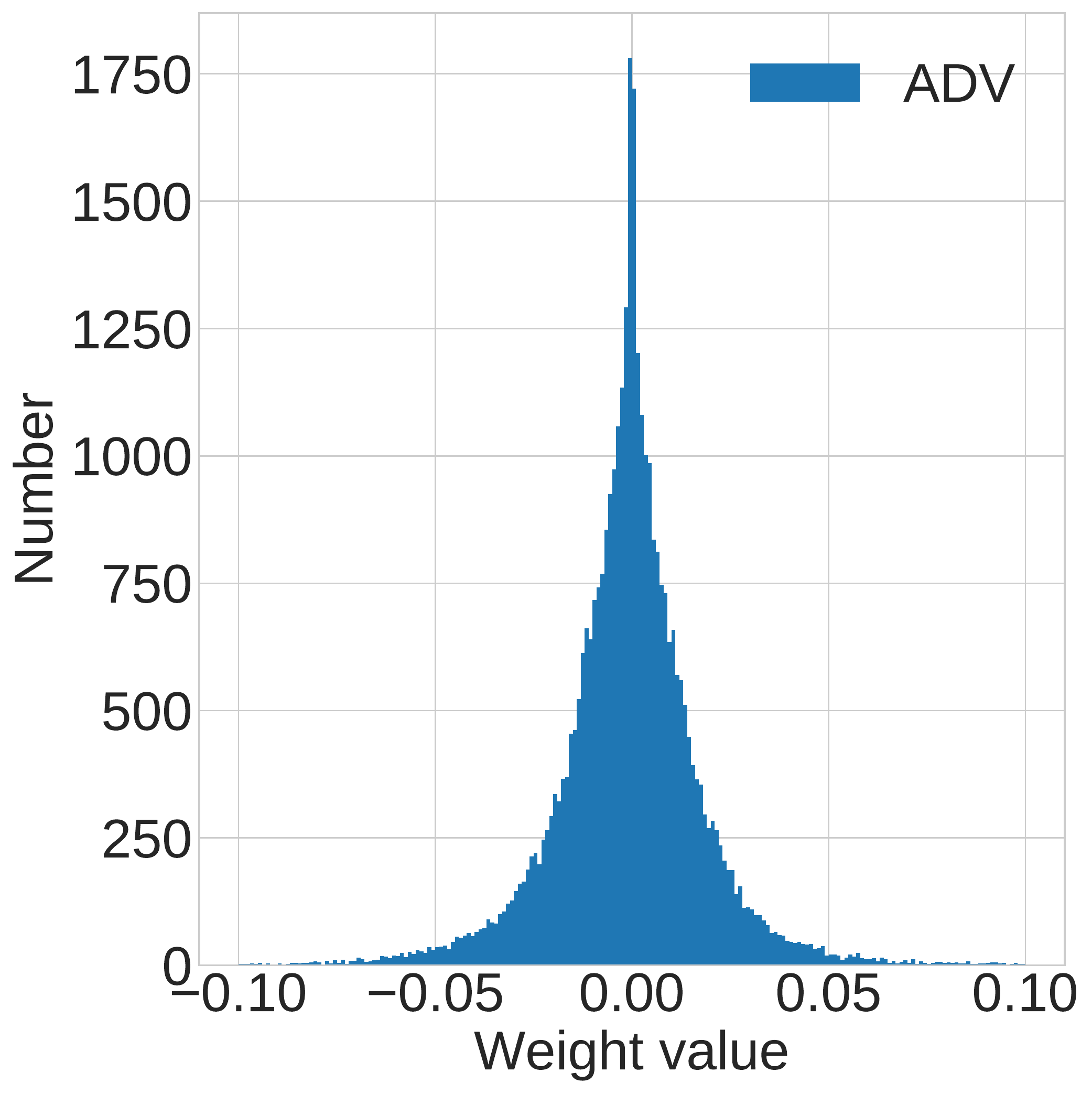} \end{minipage} \quad
\begin{minipage}[c][1\width]{0.3\textwidth}  
			\centering
			\includegraphics[width=0.9\textwidth]{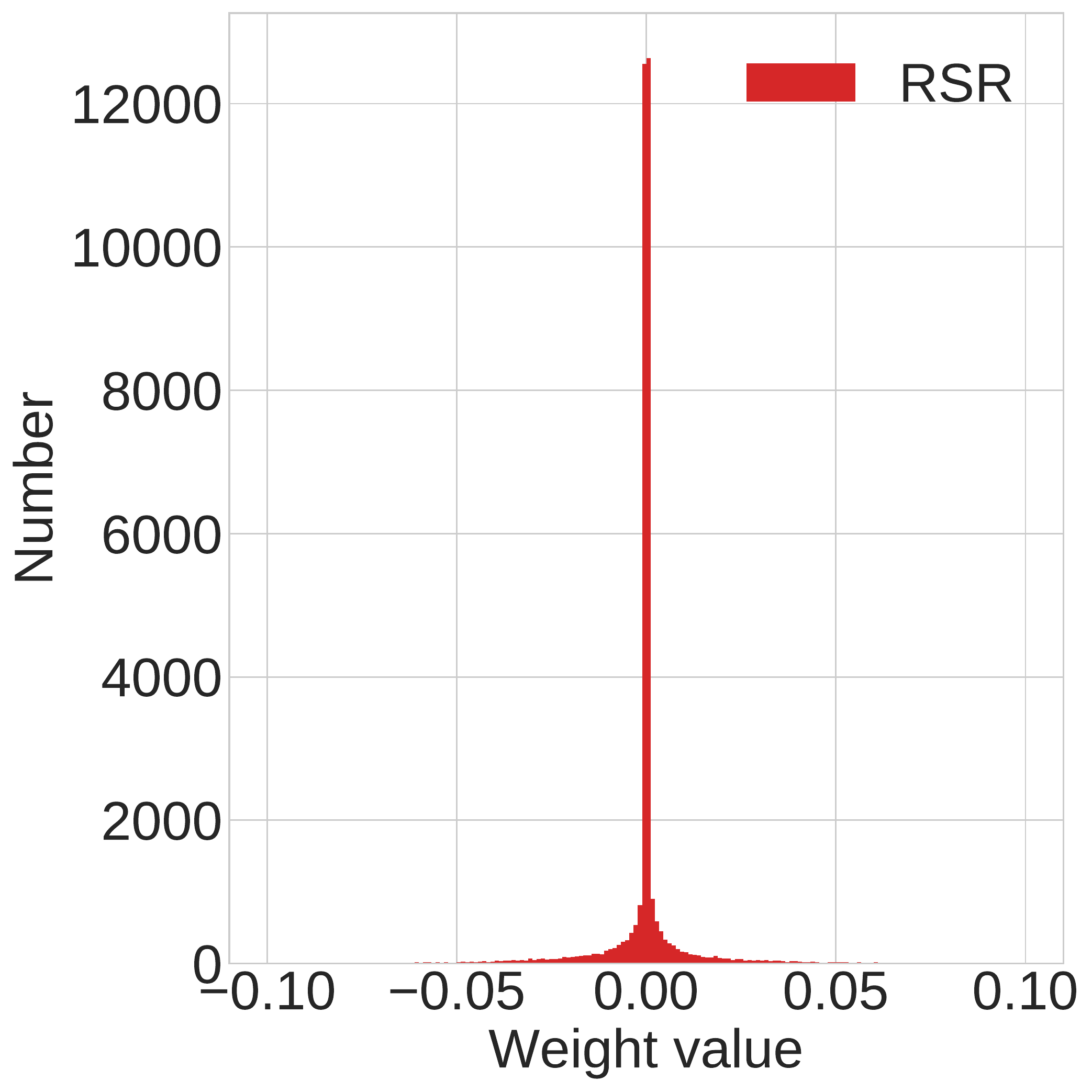} 
			\end{minipage}  
\caption{Weight distribution of ResNet-18 for the second convolution layer. We show three cases sequentially 1) clean data training 2) adversarial training(PGD) and 3) RSR training. \textbf{Observation 1}: Adversarial trained network has less sparsity than clean training thus making network pruning difficult. \textbf{Observation 2}: RSR training can achieve the desired sparseness with adversarial training.}
\label{fig:weight_dis}
\vspace{-1.5em}
\end{figure}			

\paragraph{Overview of RSR.} In this work, we propose a multi-objective optimization mechanism that could lead these two different yet related tracks, namely pursuit of network robustness and compression, to merge. To achieve this objective, we propose a novel Robust Sparse Regularization (RSR) method which integrates several regularization techniques to achieve such dual optimization. First, we propose to train a DNN with channel-wise noise injection (CNI) embedded with adversarial training to improve network robustness. Such technique injects a channel-wise Gaussian noise which is trainable during adversarial training. CNI improves test accuracy for both clean and perturbed data. Second, in order to simultaneously achieve network compactness and robustness, we propose a new ensemble loss function including an $L$-1 weight penalty term (i.e. lasso). Lasso regularization during adversarial training performs weight selection by constraining some weight values to a very small values as shown in figure \ref{fig:weight_dis}. When training is done, we could prune the small weight values based on a threshold to achieve a sparse network. Our extensive experiments show that RSR training is an effective network pruning scheme to achieve improved robustness without sacrificing any clean data accuracy across different architectures.

\section{Related Works}

\subsection{Adversarial Attack}
\label{sec:adversari}
Recent development of various powerful adversarial attack methods could totally fool a trained DNN through maliciously perturbed input data. Adversarial attacks can be generally classified into two major categories. First, white-box attack assumes the adversary has full access to the trained model and its parameters. Second, black-box attack assumes the adversary treats the model as a black box who could only access the inputs and output of the model. Now we briefly introduce various white-box and black-box attack methods which will be studied in this work.

\paragraph{FGSM Attack.}
As one of the efficient attack method, fast Gradient Sign Method (FGSM) \cite{Szegedy2013IntriguingPO} uses a single-step to generate adversarial example. If we consider a vector input $\bm{x}$ and target label $\bm{t}$, FGSM alters each element of $\bm{x}$ in the direction of its gradient w.r.t the loss $\partial \mathcal{L}/\partial x$. Similar as the original paper \cite{goodfellow2014explaining}, we define the generation of adversarial examples $\bm{\hat{x}}$ as:
\begin{equation}
\label{eqt:fgsm}
    \bm{\hat{x}} = \bm{x} + \epsilon \cdot sgn\big(\nabla_{\bm{x}}{\mathcal{L}}(g(\bm{x};\bm{\theta}),\bm{t})\big)
\end{equation}
Here, the perturbation constraint is $\epsilon$. By varying the value of $\epsilon$ we can vary the attack strength. 
$g(\bm{x};\bm{\theta})$ calculates the output of DNN with parameters $\bm{\theta}$. $sgn(\cdot)$ is the sign function. To ensure valid pixel range for the images we clip the value of $\hat{x}$ such that $\hat{x}\in [0,1]$.

\paragraph{PGD Attack.}

Projected Gradient Descent (PGD) \cite{madry2018towards} is a powerful attack with multiple steps. It is an iterative version of FGSM, $\bm{\hat{x}}^{k=1}=\bm{x}$ as the initialization. The perturbed data $\bm{\hat{x}}$ is updated during multi-step iteration process. At $k$-th step it can be expressed as:
\begin{equation}
\label{eqt:pgd}
    \bm{\hat{x}}^{k}=\Pi_{P_\epsilon(\bm{x})} \Big( \bm{\hat{x}}^{k-1} + a \cdot sgn\big(\nabla_{\bm{x}}{\mathcal{L}}(g(\bm{\hat{x}}^{k-1};\bm{\theta}),t)\big)\Big)
\end{equation}
where $P_\epsilon(\bm{x})$ is the projection space and $a$ is the step size. In \cite{madry2018towards}, it is also claimed that PGD is a universal adversary among all the first-order adversaries since it relies only on first-order information.

\paragraph{Black-box Attacks}
In this work, we also evaluate our proposed method against a wide range of Black-Box attacks. Specifically we investigate the transferable adversarial attack \cite{liu2016delving}. In transferable adversarial attack, the adversarial example is generated from source model to attack another target model. Both the source model and target model can be different but they are trained on the same training dataset. Moreover, Zero-th Order Optimization (ZOO) attack \cite{chen2017zoo} is also investigated in this work. To perform ZOO attack it does not require training a substitute model, it can directly approximate the gradient just based on the input data and output scores from the model.

\subsection{Adversarial Defenses}

Several works  \cite{madry2018towards,goodfellow2014explaining} have proposed to jointly train the network with adversarial and clean samples, called adversarial training, to achieve network robustness. Later, development of backward pass differential attack (BPDA) \cite{pmlr-v80-athalye18a} has exposed the underlying vulnerability of many other defense methods relying on gradient obfuscation \cite{s.2018stochastic,xie2018mitigating}. Then, training the network with adversarial examples has become one of the most popular defense approach to defend adversarial examples. Meanwhile, there is a cohort of work investigating the effect of regularization techniques such as quantization \cite{rakin2018defend,lin2018defensive} , noise injection \cite{he2019PNI,lecuyer2018certified,liu2017towards,yoshida2017spectral,lecuyer2018connection} and pruning \cite{s.2018stochastic,ye2019second,ye2018defending,guo2018sparse} to improve the robustness. 
Several previous works have investigated the effects of network pruning on robustness \cite{guo2018sparse,ye2019second,ye2018defending}. Recently, \cite{ye2019second} proposed concurrent weight pruning and adversarial training to generate robust and sparse network. However, their ADMM based pruning method's performance on smaller network (i.e, lesser width) suffers from poor test accuracy for both clean and adversarial data. Further, \cite{guo2018sparse} showed that pruned network will defend adversarial examples provided that the network is not over-sparsified. 

\section{Approach}

In this section, we first introduce the proposed \textit{Robust Sparse Regularization} (RSR) technique, which is incorporated into a multi-objective optimization process that simultaneously improves network robustness and compactness. Our proposed RSR mainly consists of two components: 1) a trainable Channel-wise Noise Injection (CNI) and 2) lasso weight penalty ($L$-1 norm) for model pruning, which will be introduced in this section.


\subsection{Adversarial Training}
\label{sec:adv_train}
Training the neural network with adversarial examples is a popular defense method \cite{madry2018towards,goodfellow2014explaining}. Since our method integrates with such adversarial training, we briefly introduce it first. The goal of adversarial training can be formalized as: if we have a set of inputs- $\bm{x}$ and target labels- $t$, adversarial training tries to obtain the optimal solution of network parameters $\bm{\theta}$ (i.e, weights, biases) for the following min-max optimization problem:

\begin{equation}
\label{eqt:min_max_game}
    \argmin_{\bm{\theta}} \big\{ \argmax_{\bm{x}' \in P_\epsilon(\bm{x})} \mathcal{L}\big( g(\hat{\bm{x}};\bm{\theta}), t \big) \big\}
\end{equation}

where the min-max optimization is composed of inner maximization and outer minimization problem. For inner maximization we acquire the perturbed data $\hat{\bm{x}}$ as shown in the description of PGD attack \cite{madry2018towards}. While the outer minimization is optimized through gradient descent method during network training.

\subsection{Channel-wise Noise Injection}
The first regularization technique used in RSR is to inject learnable channel-wise noise on weights during the DNN adversarial training process. Considering a convolution layer in DNN with 4-D weight tensor $\bm{W} \in \mathbb{R}^{q\times p\times kh\times kw}$, where $q,p,kh,kw$ denotes number of output channel, input channel, kernel height and kernel width respectively, the \textit{Channel-wise Noise Injection} (CNI) can be 
mathematically described as: 
\begin{equation}
\label{eqt:PNI_definition}
    \Tilde{\bm{W}} = f_\textup{CNI}(\bm{W}) = \bm{W} + \bm{\alpha} \times \bm{\eta}; \quad \bm{\eta} \sim \mathcal{N}(0, \sigma^2)
\end{equation}
where $\bm{\alpha} \in \mathbb{R}^{q,1,1,1}$ is trainable noise scaling coefficient. $\bm{\eta} \in \mathbb{R}^{q\times p\times kh\times kw}$ is the noise tensor where its elements are independently sampled from a Gaussian distributed source with zero mean and variance as $\sigma^2$. Note that, $\sigma^2$ is the variance of $\bm{W}$ that is statistically calculated in the run-time. Preliminary work \cite{he2019PNI} shows that parametric noise injection is an improved variant of adversarial training, where such trainable noise injection method could effectively regularize DNN during the adversarial training. We follow similar optimization and update rule for $\alpha$, but extending it into channel-wise version, where weights for each output channel shares the same noise scaling coefficient instead of whole layer. 


We train the network with both clean and adversarial samples to achieve a good balance between adversarial data and clean test data accuracy. Optimization problem of equation \ref{eqt:min_max_game} can be solved by minimizing the ensemble loss $\mathcal{L}_\textup{ens}$ in equation \ref{eqt:ensemble_loss_fj}.
The ensemble loss is basically the weighted sum of losses for clean and adversarial data with channel-wise trainable noise injected on weights of DNN model:
\begin{equation}
\label{eqt:ensemble_loss_fj}
   \mathcal{L}_\textup{ens}= a \cdot \mathcal{L}(g(\bm{x};f_\textup{CNI}(\bm{\theta})), \bm{t}) +  (1-a) \cdot \mathcal{L}(g(\hat{\bm{x}};f_\textup{CNI}(\bm{\theta})),\bm{t}) 
\end{equation}
where $a$ is the coefficient to balance the ensemble loss terms which is chosen as 0.5 by default. Optimizing the loss function $\mathcal{L}$ improves network robustness. The optimizer tries to solve for both model parameters $\bm{\theta}$ and $\bm{\alpha}$ to find an equilibrium between clean and perturbed data.

\subsection{Lasso Weight Penalty}
For incorporating the network pruning into the adversarial training, we propose to train the neural network with lasso weight penalty. \textit{Lasso} is known as least absolute shrinkage and selection operator \cite{tibshirani1996regression}. It was introduced as a $L$-1 regularizer that penalizes the features with higher values. Lasso is an ideal choice for weight pruning as it shrinks the lesser important weights to zero \cite{he2017channel,wang2018structured,wen2016learning}. We include the lasso weight penalty term into $\mathcal{L}_\textup{ens}$ and reformat equation \ref{eqt:ensemble_loss_fj} as:
\begin{equation}
\label{eqt:ensemble_loss_f}
   \mathcal{L}_\textup{ens}= a \cdot \mathcal{L}(g(\bm{x};f_\textup{CNI}(\bm{\theta})), t) +  (1-a)\cdot \mathcal{L}(g(\hat{\bm{x}};f_\textup{CNI}(\bm{\theta})),t) + \lambda \cdot
    \sum_{l=1}^{L}
    ||\bm{W}||_1
\end{equation}
where $\bm{W}_l$ denotes the weight tensor of $l$-th layer, and $L$ is the total number of parametric layers (i.e., convolution and fully-connected layer). $||\cdot||$ is the absolute sum of all the elements of a tensor. The effect of lasso weight penalty is determined by the coefficient $\lambda$, where $\lambda$ in larger value would generate a sparse model containing a significant amount of weight with near zero values. We tune $\lambda$ experimentally and describe the procedure for selecting optimized value of $\lambda$ in section \ref{sec:res}.

\subsection{Weight Pruning}
The proposed ensemble loss $\mathcal{L}_\textup{ens}$ serves the purpose of multi-objective loss function. We expect a network after training with the ensemble loss to be more resilient to adversarial samples. At the same time, due to the presence of lasso weight penalty, we expect a significant portion of the weight tensor to converge to near zero values. We then perform weight pruning after training with the proposed ensemble loss function, by setting the weights below a certain threshold ($\gamma$) to zero. Note that, after pruning, we remove the noise injection term for zero-value weights. As a result, during inference, we only add noise to the non-zero elements of the weight tensor. For the weight tensor in a fully connected layer, lets assume $\bm{W}=(a_{i,j})_{i,j=1}^{n,m} $ ($\bm{W} \in \mathbb{R}^{m\times n}$). For convolution layer, $\bm{W}=(a_{i,j,k,l})_{i,j,k,l=1}^{q,p,kh,kw} $ ($\bm{W} \in \mathbb{R}^{q\times p\times kh\times kw}$). Then, the pruning operation can be described as:
\begin{equation}
\label{eqt:ensemble_loss_ffs}
\resizebox{0.4\textwidth}{!}{
   FC layer $\xrightarrow{}$
   $a_{i,j}=0$ if $|a_{i,j}|< \gamma$ 
   }
\end{equation}
\begin{equation}
\label{eqt:prune}
\resizebox{0.5\textwidth}{!}{
   Conv. layer $\xrightarrow{}$
   $a_{i,j,k,l}=0$ if $|a_{i,j,k,l}|< \gamma$ 
   }
\end{equation}

Here $\gamma$ is the threshold, which is the least absolute non-zero value in the weight tensor after pruning. Again, we can tune the value of $\gamma$ for different networks to achieve different sparsity ratio. Hence, by tuning the value of $\gamma$, we can effectively show the maximum amount of parameters that can be pruned without causing robustness degradation. 

\section{Experiments}

\subsection{Experiment setup}
\paragraph{Datasets and network architectures.} 

In this work, we only consider CIFAR-10 dataset for image classification task as most of the baseline works report their robustness in terms of under-attack accuracy on this dataset. CIFAR-10 is composed of 50K training samples and 10K test samples. Our data augmentation method is same as described in \cite{he2016deep}. Attacker can directly add noise to the natural images as our data normalization layer is placed in front of the DNN as a non-trainable layer. We adopt three classical networks, ResNet-20, ResNet-18 \cite{he2016deep} and VGG-16 \cite{simonyan2014very}, to perform comparative analysis. We also show the analysis on the effect of network width by varying the width of ResNet-18 network. We report the mean accuracy with 5 trials due to the presence of randomness in both CNI and PGD \cite{madry2018towards}. We tune the hyper parameter $\lambda$ to be $1e^{-5}$ for both ResNet-18 and VGG-16 and $5e^{-5}$ for ResNet-20.  

\paragraph{Adversarial attacks.}  
In order to attack CIFAR-10 dataset using PGD attack, $\epsilon$ is set to $8/255$, and $N_\textup{step}$ is set to 7. For FGSM, attack parameters (i.e, $\epsilon$) remain the same as PGD. We use the same attack hyper parameters as in \cite{ madry2018towards}. Moreover, we also conduct the RSR defense against several state-of-the-art black-box attacks (i.e. ZOO \cite{chen2017zoo} and transfer \cite{liu2016delving} attack) in order to evaluate the proposed RSR against a wide range of attacks. 

\paragraph{Competing methods for adversarial defense.} 
In this work, PGD adversarial training \cite{madry2018towards} is selected as the primary baseline method. Additionally, our work includes channel-wise noise injection, so we also compare the method with parametric noise injection (PNI) \cite{he2019PNI}. Additionally, we also compare our work with several network compression and pruning methods \cite{ye2019second,lin2018defensive}. Finally, comparison with several state-of-the art regularization techniques serving as a adversarial defense \cite{lecuyer2018certified,liu2017towards} is also presented.

\subsection{Results}
\label{sec:res}

\paragraph{White-Box Attack.}
Our simulation results on two popular white-box attack PGD \cite{madry2018towards} and FGSM \cite{goodfellow2014explaining} are presented in table \ref{table:cifar}. During adversarial training, as stated in section \ref{sec:adv_train}, we use PGD algorithm to generate the adversarial samples. First, for the regular models, we do not perform any weight pruning. RSR helps to achieve significant robustness enhancement and even improves the clean data accuracy compared to baseline PGD training \cite{madry2018towards}. We observe that, with increasing the model capacity, network robustness increases as well. The observation of robustness enhancement with increasing network capacity is consistent with previous works \cite{madry2018towards, he2019PNI}. For our proposed RSR, the pattern remains the same. Our best accuracy was obtained using VGG-16 network. We could improve the clean test data accuracy by 0.95\% and perturbed data accuracy by 9.48\% Under strong PGD attack for VGG-16.

\begin{table*}[ht]
\begin{center}
\caption{\textbf{Summary of CIFAR-10 Results:} We report clean and perturbed-data(under PGD and FGSM attack) accuracy (\%) on CIFAR-10 test data. To visualize the effect of lasso and CNI, we also report the independent test accuracy for both channel-wise noise injection (CNI) and lasso loss. We report the percentage of weight being pruned (exactly equal to zero) as the sparsity(\%). Capacity denotes the number of trainable parameters in the network.} 

\label{table:cifar}

{
\scalebox{0.7}{
\begin{tabular}{c c c c c c c c c c c c c c c c c c}
\toprule
  &  \multicolumn{4}{c}{ResNet-20} &  \multicolumn{4}{c}{ResNet-18} & \multicolumn{4}{c}{VGG-16}  & \\
Capacity & \multicolumn{4}{c}{269,722} &  \multicolumn{4}{c}{11,173,962} & \multicolumn{4}{c}{138,357,544}  & \\ \hline
Scheme & Clean & PGD & FGSM & Sparsity & Clean & PGD & FGSM & Sparsity & Clean & PGD & FGSM & Sparsity & \\ \midrule
  &  &  &  &  &  & \textbf{Before Pruning} &    &  &   &  &  & &\\ \midrule

PGD  & 83.58 & 39.44 & 46.87 & 0 & 86.11 & 44.31 & 53.52 & 0 & 82.88 & 37.57 & 46.94 & 0 &\\ 

CNI & 84.67 & 46.11 & 54.40 & 0 & 86.82 & 47.85 & 56.04 & 0 & 83.13 & 44.23 & 51.56 & 0 &\\ 

Lasso & 83.56 & 38.69 & 45.78 & 0 & 85.92 & 46.94 & 55.2 & 0 & 83.26 & 41.93 & 50.33 & 0 &\\ 

RSR & \textbf{84.96} & \textbf{47.95} & \textbf{56.72} & 0 & \textbf{86.95} & \textbf{52.94} & \textbf{60.89} & 0 & \textbf{83.83} & \textbf{47.05} & \textbf{54.05} & 0 &\\ \midrule 
  &  &  &  &  &  & \textbf{After Pruning} &  &  &  &  &  &  &\\ \midrule
PGD & 51.58 & 12.49 & 16.11 & 60.47 & 70.31 & 31.00 & 35.8 & 85.43 & 78.40 & 32.14 & 42.21 & 50.62 &\\ 
CNI & 55.93 & 23.91 & 29.11 & 60.74 & 50.97 & 22.54 & 25.31 & 85.27 & 75.79 & 40.39 & 46.37 & 50.62 &\\ 

Lasso & 83.64 & 38.46 & 45.44 & 60.14 & 85.92 & 46.8 & 55.2 & 85.38 & 83.24 & 42.01 & 50.32 & 50.15 &\\ 

RSR & \textbf{84.32} & \textbf{47.44} & \textbf{55.74} & \textbf{60.85} & \textbf{86.79} & \textbf{53.03} & \textbf{60.35} & \textbf{85.36} & \textbf{83.02} & \textbf{47.70} & \textbf{54.16} & \textbf{50.93} &\\
\bottomrule
\end{tabular}}}
\end{center}
\end{table*}
Our proposed RSR can prune 60\%, 85\% and 50\% of the network's weight for ResNet-20, ResNet-18 and VGG-16, respectively, without any clean test accuracy loss. To show the comparative effect of network robustness and sparsity, we prune each of the four training cases (PGD/ CNI/ Lasso/ RSR) by equal amount. The level of sparsity can always be tuned by choosing different values of $\gamma$. As expected, both PGD and CNI performance suffers significantly after pruning. On the contrary, RSR outperforms baseline PGD (even without pruning) training method. We observe 8.72\% and 6.83\% improvement on test accuracy under PGD and FGSM attack respectively for ResNet-18 architecture. Again the most significant improvement observed in the VGG-16 network which has the largest capacity. Such observation confirms that increasing the number of parameters increases the effect of weight penalty and noise injection to enhance the network robustness. Another question to be asked is what if we want to prune the network beyond the reported sparsity. For example, if we want to prune ResNet-18 beyond 85\% , does the network still remain robust? We try to answer this question in the next paragraph where we explain the effect of network width with sparsity.

\paragraph{Effect of Network Width.}

\begin{figure}[ht]
    \centering

    \begin{subfigure}[t]{0.32\textwidth}
        \centering
        
        \includegraphics[height=1.2in]{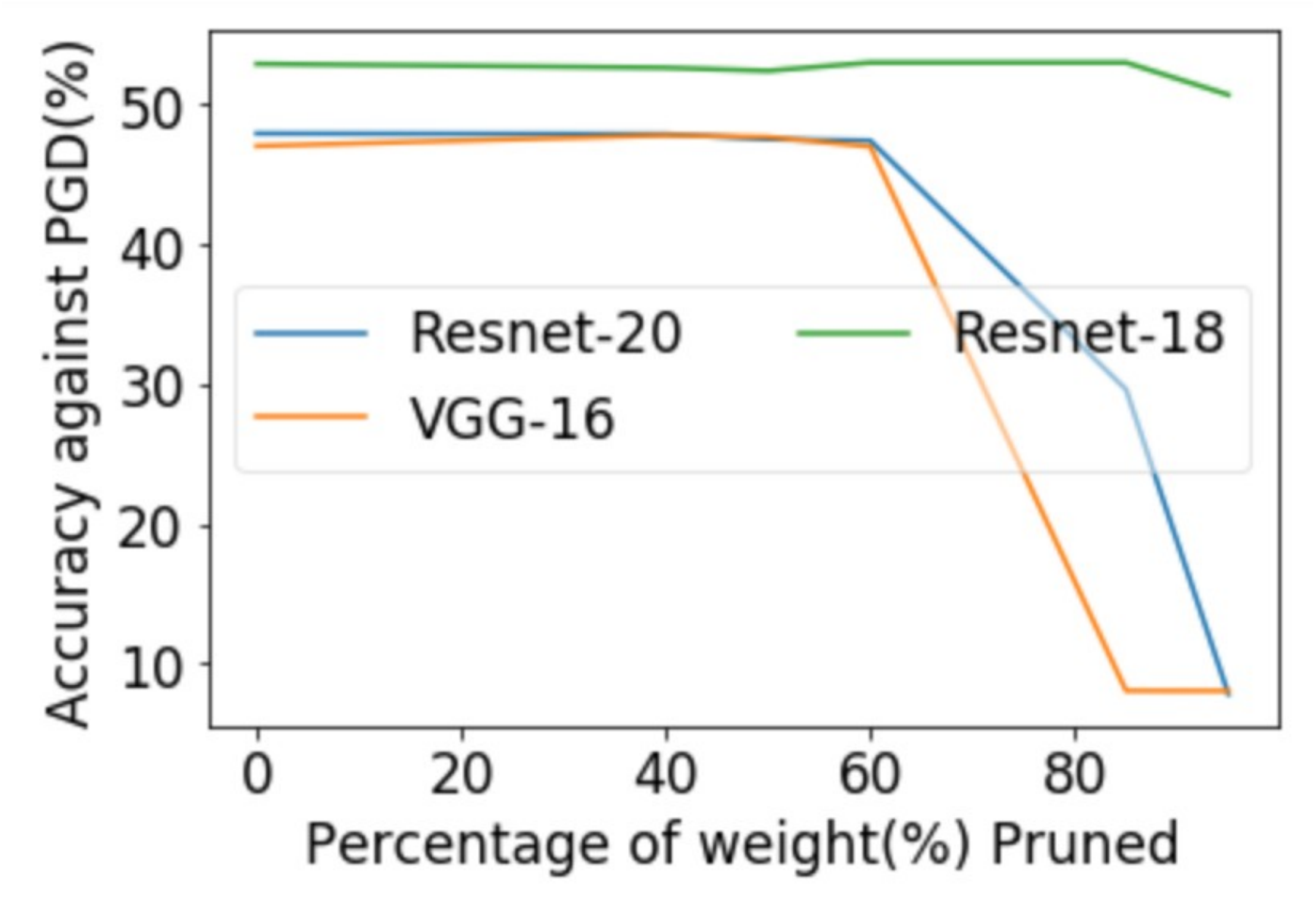}
        \caption{}
        \label{fig:sub11}
    \end{subfigure}%
    ~
    \begin{subfigure}[t]{0.32\textwidth}
        \centering
        
        \includegraphics[height=1.2in]{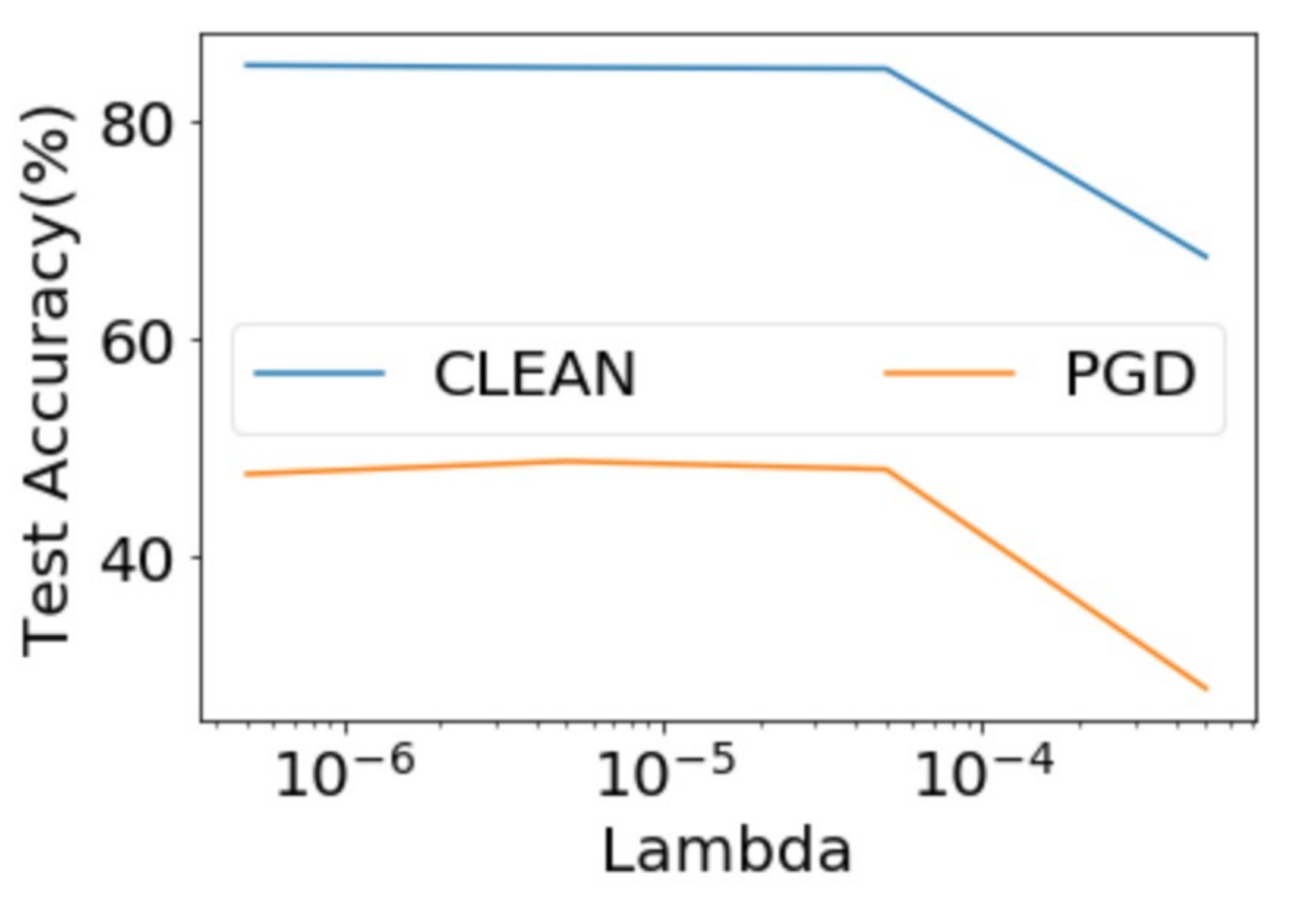}
        \caption{}
        \label{fig:sub1}
    \end{subfigure}%
    ~
    \begin{subfigure}[t]{0.32\textwidth}
        \centering
        
        \includegraphics[height=1.2in]{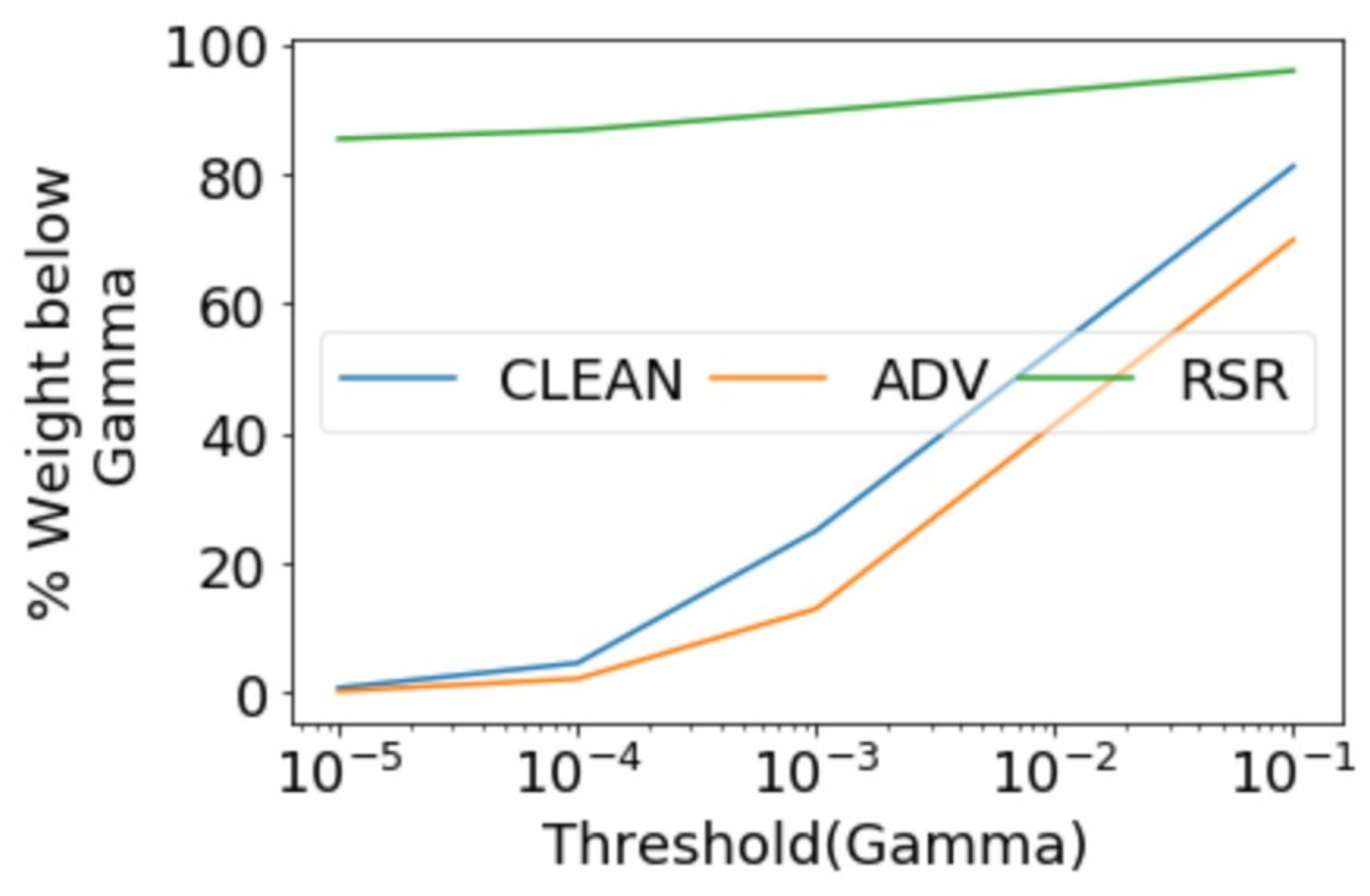}
        \caption{}
        \label{fig:sub2} 
    \end{subfigure}
    \caption{a) The relationship between test accuracy (\%) under PGD attack Vs Percentage of weight pruned(exactly equal to zero). It shows each network can be pruned up to certain level of sparsity. Pruning beyond that level would make the model over-sparsified \cite{guo2018sparse} and the network no longer remains robust. b) The plot shows both clean and perturbed data (PGD) accuracy (\%) for ResNet-20(RSR) VS Lambda($\lambda$). $\lambda$ is the regularization parameter for the lasso loss.  c)  X- axis contains different Gamma($\gamma$) values and Y-axis shows the percentage of weight below a certain threshold Gamma($\gamma$). $\gamma$ is least absolute value after pruning in a network. Clean, Adv and RSR denotes clean test data training method, adversarial training method and Our proposed RSR method respectively. This plot is only for convolution layer of ResNet-18 architecture. }
    \vspace{-1.5em}
\end{figure}

\cite{ye2019second} demonstrated that decreasing a network width may have negative impact on robustness. To verify if our method also follows the same trend, we show an ablation study on ResNet-18 with decreasing network width in table \ref{tab:width}. Our observation confirms that RSR method still remains more robust than the baseline PGD method for each case (i.e., $w=0.25\times/ 0.5\times / 1\times$). On the other side, we achieve less sparsity on network with smaller network width. In $w=0.25\times$ case, we could only achieve 38.33\% sparsity without sacrificing any clean or perturbed data accuracy. This observation is quite intuitive as ResNet-18($0.25\times$) network already has $0.125\times$ less parameters than that of ResNet-18 ($1\times$). Thus, even after performing less amount of weight pruning, the percentage of parameter (7.7\%) in the network would still be smaller compared to ResNet-18($1\times$) (14.37\%). Finally, such observation also answers the question asked previously: A particular architecture (e.g., ResNet-18) can be pruned up to a certain amount of sparsity levels based on the network width. The maximum number of parameters which can be pruned without any sacrifice in robustness may vary across different architectures. Figure \ref{fig:sub11} shows ResNet-20, ResNet-18 and VGG-16 test accuracy under PGD attack starts to drop at different sparsity levels (\% weight equal to zero). If any model is sparsified beyond this point, it falls under the definition of over-sparsified model \cite{guo2018sparse} and the network no longer remains robust.

\begin{table}[ht]
\vspace{-1em}
\centering
\caption{\textbf{Ablation study with varying width.} We report clean and perturbed-data(under PGD and FGSM attack) accuracy on CIFAR-10. ResNet-18($1\times$) is chosen as the baseline. Network width $w=0.5\times$ and $w=0.25\times$ denotes that the width of the network's both input channel and output channel is scaled by $0.5\times$ and $0.25\times$ respectively.}
\label{tab:width}
\scalebox{0.9}{
\begin{tabular}{cccccccc}
\toprule
& \multicolumn{2}{c}{\begin{tabular}[c]{@{}c@{}}Clean\\ Test (\%)\end{tabular}} & \multicolumn{2}{c}{\begin{tabular}[c]{@{}c@{}}Adversarial\\ Attack(PGD)\\ \%\end{tabular}}& \multicolumn{2}{c}{\begin{tabular}[c]{@{}c@{}}Sparsity:\\ (\%)\end{tabular}} & \begin{tabular}[c]{@{}c@{}} (\%) of parameter\\ remain in the network\\ compared to ResNet-18($1\times$)\end{tabular} \\ \hline
\begin{tabular}[c]{@{}c@{}}Channel\\ Width\end{tabular} & \begin{tabular}[c]{@{}c@{}}Adv.\\ Trained\end{tabular} & RSR & \begin{tabular}[c]{@{}c@{}}Adv. \\ Trained\end{tabular} & RSR & \begin{tabular}[c]{@{}c@{}}Adv. \\ Trained\end{tabular} & RSR & RSR \\ \hline
$w=0.25\times$ & 82.68 & 83.18 & 39.01 & 45.38 & 0 & 38.33 & $\xrightarrow{ }$(100-38.33)$\times$0.125=7.7 \\
$w=0.5\times$ & 84.99 & 84.85 & 43.33 & 50.7 & 0 & 63.17 & $\xrightarrow{ }$(100-63.17)$\times$0.25=9.21 \\
$w=1\times$ & 86.82 & 86.79 & 47.85 & 53.03 & 0 & \multicolumn{1}{c}{85.36} & $\xrightarrow{ }$(100-85.36)=14.37 \\
\bottomrule
\end{tabular}}
\end{table}

\paragraph{Robustness improvement coming from Lasso training? or CNI training? or Both?}
\label{sec:robus}
We have provided comprehensive experimental analysis on our proposed RSR method to show its performance enhancement on three fronts: clean data accuracy, robustness (i.e. under attack accuracy) and sparsity. Table \ref{table:cifar}, confirms that lasso loss primarily contributes to the sparse model generation through weight shrinkage. However, in order to identify the chief contributor towards robustness improvement, an ablation study is shown in table \ref{table:cifar}, where we also report effect of training the network only with lasso loss and CNI, respectively.  
The regularization effect of lasso is less significant for ResNet-20 and CNI plays the dominant role in network robustness improvement. However, both lasso and Channel-wise noise injection contributes towards the improvement of robustness for redundant networks (i.e, VGG-16). Both lasso and CNI can improve the network robustness by close to 4 \%  and 7 \%, respectively, on VGG-16. Nonetheless, we choose lasso because it helps shrink weight values to a very small value, thus performing a robust model selection during adversarial training. Apart from that, lasso regularization also supplements CNI towards defending adversarial examples . 

\paragraph{Choice of Lambda ($\lambda$).} In figure \ref{fig:sub1}, we show a plot of test accuracy on both clean and perturbed data versus Lambda($\lambda$) for ResNet-20. Clearly, both the test accuracy starts to drop if we increase $\lambda$ beyond $5e^{-5}$. So for ResNet-20 we choose $5e^{-5}$ as the standard value of $\lambda$ to achieve the maximum sparsity without any degradation in test accuracy. Similarly, the value of $\lambda$ for other architectures (i.e, ResNet-18, VGG-16) is optimized experimentally.

\paragraph{Black-Box Attack.}

We report the black-box attack accuracy for ResNet-20 architecture in table \ref{tab:ss}. We test our defense method against un-targeted ZOO attack \cite{chen2017zoo}. We randomly select 200 test samples to calculate the attack success rate. Our proposed method defends ZOO attack better as it decreases the attack success rate by 12 \% compared to baseline PGD method.
\begin{table}[ht]
\vspace{-1em}
\centering
\caption{\textbf{Black-Box attack summary.}  ZOO attack success rate (in 2\textsuperscript{nd} column) is the percentage of test sample being successfully classified to a wrong class by the attack. We report two sets of transfer attack accuracy: one with VGG16 as the source (3\textsuperscript{rd} column) and the other with ResNet-18 as the source (4\textsuperscript{th} column). For both PGD and RSR ResNet-20 is the target model.}
\label{tab:ss}
\scalebox{0.9}{\begin{tabular}{cccc}
\toprule
Method & \begin{tabular}[c]{@{}c@{}}ZOO Success \\  rate (\%)\end{tabular} & \begin{tabular}[c]{@{}c@{}}Source(VGG-16) \\ Accuracy(\%)\end{tabular} & \begin{tabular}[c]{@{}c@{}}Source(ResNet-18)\\ Accuracy(\%)\end{tabular} \\ \midrule
PGD & 68.50 & 66.13 & 67.44 \\
RSR & \textbf{56.00} & 66.04 & 67.27 \\
\bottomrule
\end{tabular}}
\end{table}

To perform transferable attack on RSR and PGD, we use VGG-16 and ResNet-18 network as the source model. For both cases, our RSR performs on par with the PGD method. Additionally, our proposed RSR reports higher test accuracy against black-box attack compared to white-box PGD method. Better resistance against black-box attack is considered as a sign of a defense that does not effectively uses obfuscated or masked gradient \cite{pmlr-v80-athalye18a} as a defense tool.


\paragraph{Comparison to state-of-the art techniques.}
In table \ref{tab:compare}, we summarize the performance of our defense in comparison to some other state-of-the-art defense techniques. Our proposed RSR method outperforms these comparative defenses and achieves significant robustness improvement. 

\begin{table}[ht]
\vspace{-1em}
\centering
\caption{We compare our method with three major categories of defense: a) Adversarial training defenses: Projected Gradient Descent (PGD) training \cite{madry2018towards}, Parametric Noise Injection (PNI) \cite{he2019PNI} b) Compression or pruning techniques: Defensive Quantization (DQ) \cite{lin2018defensive}, Second Rethinking of Network Pruning (SR) \cite{ye2019second} and c) Regularization techniques: Differential Privacy (DP) \cite{lecuyer2018certified} and Robust Self Ensemble(RSE) \cite{liu2017towards}.}
\label{tab:compare}
\scalebox{0.8}{\begin{tabular}{cccccccc}
\toprule
& \multicolumn{2}{c}{\textbf{\begin{tabular}[c]{@{}c@{}}Adversarial \\ Training\end{tabular}}} & \multicolumn{2}{c}{\textbf{Compression}} & \multicolumn{2}{c}{\textbf{Regularization}} & \textbf{This work} \\ \cline{2-8} 
 & PGD & PNI & DQ & SR & DP & RSE & RSR \\ \midrule
Model & ResNet-18 & ResNet-20(4$\times$) & Wide ResNet & ResNet-18 & Wide ResNet & ResNext & ResNet-18 \\
Clean(\%) & 86.11 & \textbf{87.7} & 87.0 & 81.83 & 87.0 & 87.5 & 86.79 \\
PGD(\%) & 44.31 & 49.1 & 51.8 & 48.00 & 25.0 & 40.0 & \textbf{53.03} \\
Sparsity (\%) & 0 & 0 & (6$\times$) Compression & - & 0 & 0 & \textbf{85.36} \\
\bottomrule
\end{tabular}}
\end{table}
Note that, we compare with the unbroken defenses that are not reported to show signs of obfuscated gradients \cite{pmlr-v80-athalye18a} yet. Again there are some previous works on network pruning and robustness \cite{s.2018stochastic} which might suffer from gradient obfuscation \cite{pmlr-v80-athalye18a}. \cite{guo2018sparse} first theoretically shows the effect of pruning on non-linear DNN to demonstrate the vulnerability of over-sparsified model to adversarial attacks. However, we are the first to formulate an improved adversarial defense with sparse regularization. Our proposed RSR generates sparse and compact neural network that can achieve state-of-the-art under-attack accuracy and much improved robustness. 

\vspace{-1em}
\section{Analysis}
\label{sec:ana}
\paragraph{RSR is performing regularization.}
Robust Sparse Regularization is performing regularization on the network to enhance both robustness and compactness. It does not show any obvious signs of gradient masking proposed in \cite{pmlr-v80-athalye18a}. First, RSR performs better against single step attack (i.e, FGSM) compared to multiple step attack (i.e., PGD). Also we report higher test accuracy against black-box attack than white-box. Finally, increasing the attack strength linearly decreases the effectiveness of our defense. Such observations confirm primarily our robustness enhancement is not achieved through any gradient obfuscation or masking \cite{pmlr-v80-athalye18a}. Instead, our improvement in robustness primarily comes from regularized training method. However, after pruning , presence of noise in the surviving weights is also playing a critical role. Table \ref{tab:analysis} summarizes the impact of the presence of noise during the inference on robustness.

\begin{table}[ht]
\vspace{-1em}
\centering
\caption{We report the clean and perturbed data (PGD) test accuracy for two cases 1) with inference noise and 2) without inference noise. We choose VGG-16 with 50 \% sparsity.}
\label{tab:analysis}
\begin{tabular}{ccccccc}
\toprule
 & \multicolumn{2}{c}{With inference Noise} & \multicolumn{2}{c}{Without Inference Noise} & \multicolumn{2}{c}{Baseline(PGD)} \\ 
\multicolumn{1}{c}{} & Clean & PGD & Clean & PGD & Clean & PGD \\ \midrule
RSR & 83.02 & 47.70 & 83.47 & 41.61 & 82.88 & 37.57 \\
\bottomrule
\end{tabular}
\end{table}

When we disable the inference noise in the network, test accuracy under PGD attack drops to 41.61 \%. This confirms the presence of inference noise contributing heavily towards the robustness. However, our regularization training still stands out as even after the accuracy drop we maintain a higher accuracy than the baseline. Thus we conclude the robustness achieved in this work can be viewed as a combined effect of regularization(i.e., CNI and lasso), sparsity and inference noise.

\paragraph{Optimal Gamma provides the improvement on three fronts.}
We can fine-tune the model after training to prune the weights of a network below a certain threshold ($\gamma$). During training apart from enhancing robustness, RSR mainly shrinks the weights of the network. The demonstration of weight shrinkage is presented in figure \ref{fig:sub2}. ResNet-18 network training with RSR contains 85\% weights with near zero value (less than $1e^{-5}$). So pruning weights with such small values will have minimal effect on clean test accuracy and robustness. Thus, the value of $\gamma$ can be tuned to an optimal point for each network to achieve improvement on three fronts: clean data accuracy, robustness(i.e., under attack accuracy) and sparsity.

\section{Conclusion}
We successfully co-optimize the objective of network robustness and compactness through our proposed RSR training method. As a result, we show that heavily sparse network can resist adversarial examples to generate both robust and compact neural network at the same time. Our proposed method performs dual optimization during training to resist state-of-art white-box and black-box attacks using a more compact network.

{
\small
\bibliography{egbib}

@article{bietti2018regularization,
  title={On Regularization and Robustness of Deep Neural Networks},
  author={Bietti, Alberto and Mialon, Gr{\'e}goire and Mairal, Julien},
  journal={arXiv preprint arXiv:1810.00363},
  year={2018}
}
@article{liu2017towards,
  title={Towards Robust Neural Networks via Random Self-ensemble},
  author={Liu, Xuanqing and Cheng, Minhao and Zhang, Huan and Hsieh, Cho-Jui},
  journal={arXiv preprint arXiv:1712.00673},
  year={2017}
}
@article{goodfellow2014explaining,
  title={Explaining and harnessing adversarial examples},
  author={Goodfellow, Ian J and Shlens, Jonathon and Szegedy, Christian},
  journal={arXiv preprint arXiv:1412.6572},
  year={2014}
}
@article{athalye2018obfuscated,
  title={Obfuscated gradients give a false sense of security: Circumventing defenses to adversarial examples},
  author={Athalye, Anish and Carlini, Nicholas and Wagner, David},
  journal={arXiv preprint arXiv:1802.00420},
  year={2018}
}







@InProceedings{pmlr-v80-athalye18a,
  title = 	 {Obfuscated Gradients Give a False Sense of Security: Circumventing Defenses to Adversarial Examples},
  author = 	 {Athalye, Anish and Carlini, Nicholas and Wagner, David},
  booktitle = 	 {Proceedings of the 35th International Conference on Machine Learning},
  pages = 	 {274--283},
  year = 	 {2018},
  editor = 	 {Dy, Jennifer and Krause, Andreas},
  volume = 	 {80},
  series = 	 {Proceedings of Machine Learning Research},
  address = 	 {Stockholmsmässan, Stockholm Sweden},
  month = 	 {10--15 Jul},
  publisher = 	 {PMLR},
  pdf = 	 {http://proceedings.mlr.press/v80/athalye18a/athalye18a.pdf},
  url = 	 {http://proceedings.mlr.press/v80/athalye18a.html},
  abstract = 	 {We identify obfuscated gradients, a kind of gradient masking, as a phenomenon that leads to a false sense of security in defenses against adversarial examples. While defenses that cause obfuscated gradients appear to defeat iterative optimization-based attacks, we find defenses relying on this effect can be circumvented. We describe characteristic behaviors of defenses exhibiting the effect, and for each of the three types of obfuscated gradients we discover, we develop attack techniques to overcome it. In a case study, examining non-certified white-box-secure defenses at ICLR 2018, we find obfuscated gradients are a common occurrence, with 7 of 9 defenses relying on obfuscated gradients. Our new attacks successfully circumvent 6 completely, and 1 partially, in the original threat model each paper considers.}
}

@article{lecuyer2018connection,
  title={On the Connection between Differential Privacy and Adversarial Robustness in Machine Learning},
  author={Lecuyer, Mathias and Atlidakis, Vaggelis and Geambasu, Roxana and Hsu, Daniel and Jana, Suman},
  journal={arXiv preprint arXiv:1802.03471},
  year={2018}
}



@inproceedings{he2015delving,
  title={Delving deep into rectifiers: Surpassing human-level performance on imagenet classification},
  author={He, Kaiming and Zhang, Xiangyu and Ren, Shaoqing and Sun, Jian},
  booktitle={Proceedings of the IEEE international conference on computer vision},
  pages={1026--1034},
  year={2015}
}

@article{cheng2018seq2sick,
  title={Seq2Sick: Evaluating the Robustness of Sequence-to-Sequence Models with Adversarial Examples},
  author={Cheng, Minhao and Yi, Jinfeng and Zhang, Huan and Chen, Pin-Yu and Hsieh, Cho-Jui},
  journal={arXiv preprint arXiv:1803.01128},
  year={2018}
}


@article{liu2016delving,
  title={Delving into transferable adversarial examples and black-box attacks},
  author={Liu, Yanpei and Chen, Xinyun and Liu, Chang and Song, Dawn},
  journal={arXiv preprint arXiv:1611.02770},
  year={2016}
}

@inproceedings{chen2017zoo,
  title={Zoo: Zeroth order optimization based black-box attacks to deep neural networks without training substitute models},
  author={Chen, Pin-Yu and Zhang, Huan and Sharma, Yash and Yi, Jinfeng and Hsieh, Cho-Jui},
  booktitle={Proceedings of the 10th ACM Workshop on Artificial Intelligence and Security},
  pages={15--26},
  year={2017},
  organization={ACM}
}

@inproceedings{hung2017comparing,
  title={Comparing deep neural network and other machine learning algorithms for stroke prediction in a large-scale population-based electronic medical claims database},
  author={Hung, Chen-Ying and Chen, Wei-Chen and Lai, Po-Tsun and Lin, Ching-Heng and Lee, Chi-Chun},
  booktitle={2017 39th Annual International Conference of the IEEE Engineering in Medicine and Biology Society (EMBC)},
  pages={3110--3113},
  year={2017},
  organization={IEEE}
}

@incollection{NIPS2013_5025,
title = {Predicting Parameters in Deep Learning},
author = {Denil, Misha and Shakibi, Babak and Dinh, Laurent and Ranzato, Marc\textquotesingle Aurelio and de Freitas, Nando},
booktitle = {Advances in Neural Information Processing Systems 26},
editor = {C. J. C. Burges and L. Bottou and M. Welling and Z. Ghahramani and K. Q. Weinberger},
pages = {2148--2156},
year = {2013},
publisher = {Curran Associates, Inc.},
url = {http://papers.nips.cc/paper/5025-predicting-parameters-in-deep-learning.pdf}
}

@inproceedings{han2015learning,
  title={Learning both weights and connections for efficient neural network},
  author={Han, Song and Pool, Jeff and Tran, John and Dally, William},
  booktitle={Advances in neural information processing systems},
  pages={1135--1143},
  year={2015}
}

@inproceedings{molchanov2017variational,
  title={Variational dropout sparsifies deep neural networks},
  author={Molchanov, Dmitry and Ashukha, Arsenii and Vetrov, Dmitry},
  booktitle={Proceedings of the 34th International Conference on Machine Learning-Volume 70},
  pages={2498--2507},
  year={2017},
  organization={JMLR. org}
}

@inproceedings{han2016mcdnn,
  title={Mcdnn: An approximation-based execution framework for deep stream processing under resource constraints},
  author={Han, Seungyeop and Shen, Haichen and Philipose, Matthai and Agarwal, Sharad and Wolman, Alec and Krishnamurthy, Arvind},
  booktitle={Proceedings of the 14th Annual International Conference on Mobile Systems, Applications, and Services},
  pages={123--136},
  year={2016},
  organization={ACM}
}

@article{papernot2016transferability,
  title={Transferability in machine learning: from phenomena to black-box attacks using adversarial samples},
  author={Papernot, Nicolas and McDaniel, Patrick and Goodfellow, Ian},
  journal={arXiv preprint arXiv:1605.07277},
  year={2016}
}



@article{carlini2018audio,
  title={Audio Adversarial Examples: Targeted Attacks on Speech-to-Text},
  author={Carlini, Nicholas and Wagner, David},
  journal={arXiv preprint arXiv:1801.01944},
  year={2018}
}

@article{sun2018identify,
  title={Identify Susceptible Locations in Medical Records via Adversarial Attacks on Deep Predictive Models},
  author={Sun, Mengying and Tang, Fengyi and Yi, Jinfeng and Wang, Fei and Zhou, Jiayu},
  journal={arXiv preprint arXiv:1802.04822},
  year={2018}
}

@article{hinton2012deep,
  title={Deep neural networks for acoustic modeling in speech recognition: The shared views of four research groups},
  author={Hinton, Geoffrey and Deng, Li and Yu, Dong and Dahl, George E and Mohamed, Abdel-rahman and Jaitly, Navdeep and Senior, Andrew and Vanhoucke, Vincent and Nguyen, Patrick and Sainath, Tara N and others},
  journal={IEEE Signal Processing Magazine},
  volume={29},
  number={6},
  pages={82--97},
  year={2012},
  publisher={IEEE}
}

@inproceedings{ren2015faster,
  title={Faster r-cnn: Towards real-time object detection with region proposal networks},
  author={Ren, Shaoqing and He, Kaiming and Girshick, Ross and Sun, Jian},
  booktitle={Advances in neural information processing systems},
  pages={91--99},
  year={2015}
}

@article{simonyan2014very,
  title={Very deep convolutional networks for large-scale image recognition},
  author={Simonyan, Karen and Zisserman, Andrew},
  journal={arXiv preprint arXiv:1409.1556},
  year={2014}
}

@inproceedings{szegedy2017inception,
  title={Inception-v4, inception-resnet and the impact of residual connections on learning.},
  author={Szegedy, Christian and Ioffe, Sergey and Vanhoucke, Vincent and Alemi, Alexander A},
  booktitle={AAAI},
  volume={4},
  pages={12},
  year={2017}
}

@inproceedings{he2016deep,
  title={Deep residual learning for image recognition},
  author={He, Kaiming and Zhang, Xiangyu and Ren, Shaoqing and Sun, Jian},
  booktitle={Proceedings of the IEEE conference on computer vision and pattern recognition},
  pages={770--778},
  year={2016}
}

@inproceedings{krizhevsky2012imagenet,
  title={Imagenet classification with deep convolutional neural networks},
  author={Krizhevsky, Alex and Sutskever, Ilya and Hinton, Geoffrey E},
  booktitle={Advances in neural information processing systems},
  pages={1097--1105},
  year={2012}
}

@article{xiong2016achieving,
  title={Achieving human parity in conversational speech recognition},
  author={Xiong, Wayne and Droppo, Jasha and Huang, Xuedong and Seide, Frank and Seltzer, Mike and Stolcke, Andreas and Yu, Dong and Zweig, Geoffrey},
  journal={arXiv preprint arXiv:1610.05256},
  year={2016}
}
@inproceedings{biggio2013evasion,
  title={Evasion attacks against machine learning at test time},
  author={Biggio, Battista and Corona, Igino and Maiorca, Davide and Nelson, Blaine and {\v{S}}rndi{\'c}, Nedim and Laskov, Pavel and Giacinto, Giorgio and Roli, Fabio},
  booktitle={Joint European Conference on Machine Learning and Knowledge Discovery in Databases},
  pages={387--402},
  year={2013},
  organization={Springer}
}



@article{Szegedy2013IntriguingPO,
  title={Intriguing properties of neural networks},
  author={Christian Szegedy and Wojciech Zaremba and Ilya Sutskever and Joan Bruna and Dumitru Erhan and Ian J. Goodfellow and Rob Fergus},
  journal={CoRR},
  year={2013},
  volume={abs/1312.6199}
}


@article{kurakin2016adversarial,
  title={Adversarial examples in the physical world},
  author={Kurakin, Alexey and Goodfellow, Ian and Bengio, Samy},
  journal={arXiv preprint arXiv:1607.02533},
  year={2016}
}

@inproceedings{papernot2016limitations,
  title={The limitations of deep learning in adversarial settings},
  author={Papernot, Nicolas and McDaniel, Patrick and Jha, Somesh and Fredrikson, Matt and Celik, Z Berkay and Swami, Ananthram},
  booktitle={Security and Privacy (EuroS\&P), 2016 IEEE European Symposium on},
  pages={372--387},
  year={2016},
  organization={IEEE}
}

@inproceedings{moosavi2016deepfool,
  title={Deepfool: a simple and accurate method to fool deep neural networks},
  author={Moosavi-Dezfooli, Seyed-Mohsen and Fawzi, Alhussein and Frossard, Pascal},
  booktitle={Proceedings of the IEEE Conference on Computer Vision and Pattern Recognition},
  pages={2574--2582},
  year={2016}
}

@inproceedings{
madry2018towards,
title={Towards Deep Learning Models Resistant to Adversarial Attacks},
author={Aleksander Madry and Aleksandar Makelov and Ludwig Schmidt and Dimitris Tsipras and Adrian Vladu},
booktitle={International Conference on Learning Representations},
year={2018},
url={https://openreview.net/forum?id=rJzIBfZAb},
}



@article{Gu2014TowardsDN,
  title={Towards Deep Neural Network Architectures Robust to Adversarial Examples},
  author={Shixiang Gu and Luca Rigazio},
  journal={CoRR},
  year={2014},
  volume={abs/1412.5068}
}



@article{Xu2018FeatureSD,
  title={Feature Squeezing: Detecting Adversarial Examples in Deep Neural Networks},
  author={Weilin Xu and David Evans and Yanjun Qi},
  journal={CoRR},
  year={2018},
  volume={abs/1704.01155}
}



@inproceedings{
tramèr2018ensemble,
title={Ensemble Adversarial Training: Attacks and Defenses},
author={Florian Tramèr and Alexey Kurakin and Nicolas Papernot and Ian Goodfellow and Dan Boneh and Patrick McDaniel},
booktitle={International Conference on Learning Representations},
year={2018},
url={https://openreview.net/forum?id=rkZvSe-RZ},
}

@article{Goodfellow2014ExplainingAH,
  title={Explaining and Harnessing Adversarial Examples},
  author={Ian J. Goodfellow and Jonathon Shlens and Christian Szegedy},
  journal={CoRR},
  year={2014},
  volume={abs/1412.6572}
}

@article{kos2017adversarial,
  title={Adversarial examples for generative models},
  author={Kos, Jernej and Fischer, Ian and Song, Dawn},
  journal={arXiv preprint arXiv:1702.06832},
  year={2017}
}

@article{chen2017show,
  title={Show-and-Fool: Crafting Adversarial Examples for Neural Image Captioning},
  author={Chen, Hongge and Zhang, Huan and Chen, Pin-Yu and Yi, Jinfeng and Hsieh, Cho-Jui},
  journal={arXiv preprint arXiv:1712.02051},
  year={2017}
}

@article{kos2017delving,
  title={Delving into adversarial attacks on deep policies},
  author={Kos, Jernej and Song, Dawn},
  journal={arXiv preprint arXiv:1705.06452},
  year={2017}
}

@article{papernot2016towards,
  title={Towards the science of security and privacy in machine learning},
  author={Papernot, Nicolas and McDaniel, Patrick and Sinha, Arunesh and Wellman, Michael},
  journal={arXiv preprint arXiv:1611.03814},
  year={2016}
}

@inproceedings{papernot2016distillation,
  title={Distillation as a defense to adversarial perturbations against deep neural networks},
  author={Papernot, Nicolas and McDaniel, Patrick and Wu, Xi and Jha, Somesh and Swami, Ananthram},
  booktitle={Security and Privacy (SP), 2016 IEEE Symposium on},
  pages={582--597},
  year={2016},
  organization={IEEE}
}




@article{Carlini2016DefensiveDI,
  title={Defensive Distillation is Not Robust to Adversarial Examples},
  author={Nicholas Carlini and David A. Wagner},
  journal={CoRR},
  year={2016},
  volume={abs/1607.04311}
}

@inproceedings{papernot2017practical,
  title={Practical black-box attacks against machine learning},
  author={Papernot, Nicolas and McDaniel, Patrick and Goodfellow, Ian and Jha, Somesh and Celik, Z Berkay and Swami, Ananthram},
  booktitle={Proceedings of the 2017 ACM on Asia Conference on Computer and Communications Security},
  pages={506--519},
  year={2017},
  organization={ACM}
}

@article{bhagoji2017dimensionality,
  title={Dimensionality Reduction as a Defense against Evasion Attacks on Machine Learning Classifiers},
  author={Bhagoji, Arjun Nitin and Cullina, Daniel and Mittal, Prateek},
  journal={arXiv preprint arXiv:1704.02654},
  year={2017}
}

@article{Bhagoji2018EnhancingRO,
  title={Enhancing robustness of machine learning systems via data transformations},
  author={Arjun Nitin Bhagoji and Daniel Cullina and Chawin Sitawarin and Prateek Mittal},
  journal={2018 52nd Annual Conference on Information Sciences and Systems (CISS)},
  year={2018},
  pages={1-5}
}

@InProceedings{pmlr-v48-oord16,
  title = 	 {Pixel Recurrent Neural Networks},
  author = 	 {Aaron Van Oord and Nal Kalchbrenner and Koray Kavukcuoglu},
  booktitle = 	 {Proceedings of The 33rd International Conference on Machine Learning},
  pages = 	 {1747--1756},
  year = 	 {2016},
  editor = 	 {Maria Florina Balcan and Kilian Q. Weinberger},
  volume = 	 {48},
  series = 	 {Proceedings of Machine Learning Research},
  address = 	 {New York, New York, USA},
  month = 	 {20--22 Jun},
  publisher = 	 {PMLR},
  pdf = 	 {http://proceedings.mlr.press/v48/oord16.pdf},
  url = 	 {http://proceedings.mlr.press/v48/oord16.html},
  abstract = 	 {Modeling the distribution of natural images is a landmark problem in unsupervised learning. This task requires an image model that is at once expressive, tractable and scalable. We present a deep neural network that sequentially predicts the pixels in an image along the two spatial dimensions. Our method models the discrete probability of the raw pixel values and encodes the complete set of dependencies in the image. Architectural novelties include fast two-dimensional recurrent layers and an effective use of residual connections in deep recurrent networks. We achieve log-likelihood scores on natural images that are considerably better than the previous state of the art. Our main results also provide benchmarks on the diverse ImageNet dataset. Samples generated from the model appear crisp, varied and globally coherent.}
}


@article{
buckman2018thermometer,
title={Thermometer Encoding: One Hot Way To Resist Adversarial Examples},
author={Jacob Buckman, Aurko Roy, Colin Raffel, Ian Goodfellow},
journal={International Conference on Learning Representations},
year={2018},
url={https://openreview.net/forum?id=S18Su--CW},
note={accepted as poster},
}


@article{lin2017does,
  title={Why does deep and cheap learning work so well?},
  author={Lin, Henry W and Tegmark, Max and Rolnick, David},
  journal={Journal of Statistical Physics},
  volume={168},
  number={6},
  pages={1223--1247},
  year={2017},
  publisher={Springer}
}

@article{qian2018l2,
  title={L2-Nonexpansive Neural Networks},
  author={Qian, Haifeng and Wegman, Mark N},
  journal={arXiv preprint arXiv:1802.07896},
  year={2018}
}

@inproceedings{huang2017densely,
  title={Densely connected convolutional networks},
  author={Huang, Gao and Liu, Zhuang and Weinberger, Kilian Q and van der Maaten, Laurens},
  booktitle={Proceedings of the IEEE conference on computer vision and pattern recognition},
  volume={1},
  pages={3},
  year={2017}
}




@article{Ilyas2017TheRM,
  title={The Robust Manifold Defense: Adversarial Training using Generative Models},
  author={Andrew Ilyas and Ajil Jalal and Eirini Asteri and Constantinos Daskalakis and Alexandros G. Dimakis},
  journal={CoRR},
  year={2017},
  volume={abs/1712.09196}
}


@inproceedings{
guo2018countering,
title={Countering Adversarial Images using Input Transformations},
author={Chuan Guo and Mayank Rana and Moustapha Cisse and Laurens van der Maaten},
booktitle={International Conference on Learning Representations},
year={2018},
url={https://openreview.net/forum?id=SyJ7ClWCb},
}

@article{ma2018characterizing,
  title={Characterizing Adversarial Subspaces Using Local Intrinsic Dimensionality},
  author={Ma, Xingjun and Li, Bo and Wang, Yisen and Erfani, Sarah M and Wijewickrema, Sudanthi and Houle, Michael E and Schoenebeck, Grant and Song, Dawn and Bailey, James},
  journal={arXiv preprint arXiv:1801.02613},
  year={2018}
}

@inproceedings{
song2018pixeldefend,
title={PixelDefend: Leveraging Generative Models to Understand and Defend against Adversarial Examples},
author={Yang Song and Taesup Kim and Sebastian Nowozin and Stefano Ermon and Nate Kushman},
booktitle={International Conference on Learning Representations},
year={2018},
url={https://openreview.net/forum?id=rJUYGxbCW},
}



@inproceedings{
xie2018mitigating,
title={Mitigating Adversarial Effects Through Randomization},
author={Cihang Xie and Jianyu Wang and Zhishuai Zhang and Zhou Ren and Alan Yuille},
booktitle={International Conference on Learning Representations},
year={2018},
url={https://openreview.net/forum?id=Sk9yuql0Z},
}


@inproceedings{rajendran2012security,
  title={Security analysis of logic obfuscation},
  author={Rajendran, Jeyavijayan and Pino, Youngok and Sinanoglu, Ozgur and Karri, Ramesh},
  booktitle={Proceedings of the 49th Annual Design Automation Conference},
  pages={83--89},
  year={2012},
  organization={ACM}
}

@article{guin2013counterfeit,
  title={Counterfeit IC detection and challenges ahead},
  author={Guin, Ujjwal and Tehranipoor, Mohammad and DiMase, Dan and Megrdichian, Mike and others},
  journal={ACM SIGDA},
  volume={43},
  number={3},
  pages={1--5},
  year={2013}
}

@inproceedings{ioffe2015batch,
  title={Batch normalization: Accelerating deep network training by reducing internal covariate shift},
  author={Ioffe, Sergey and Szegedy, Christian},
  booktitle={International conference on machine learning},
  pages={448--456},
  year={2015}
}

@article{marques2017distributed,
  title={Distributed learning of CNNs on heterogeneous CPU/GPU architectures},
  author={Marques, Jose and Falcao, Gabriel and Alexandre, Lu{\'\i}s A},
  journal={arXiv preprint arXiv:1712.02546},
  year={2017}
}

@inproceedings{zhang2015optimizing,
  title={Optimizing fpga-based accelerator design for deep convolutional neural networks},
  author={Zhang, Chen and Li, Peng and Sun, Guangyu and Guan, Yijin and Xiao, Bingjun and Cong, Jason},
  booktitle={Proceedings of the 2015 ACM/SIGDA International Symposium on Field-Programmable Gate Arrays},
  pages={161--170},
  year={2015},
  organization={ACM}
}

@article{wu2016google,
  title={Google's neural machine translation system: Bridging the gap between human and machine translation},
  author={Wu, Yonghui and Schuster, Mike and Chen, Zhifeng and Le, Quoc V and Norouzi, Mohammad and Macherey, Wolfgang and Krikun, Maxim and Cao, Yuan and Gao, Qin and Macherey, Klaus and others},
  journal={arXiv preprint arXiv:1609.08144},
  year={2016}
}


@inproceedings{chakraborty2009hardware,
  title={Hardware Trojan: Threats and emerging solutions},
  author={Chakraborty, Rajat Subhra and Narasimhan, Seetharam and Bhunia, Swarup},
  booktitle={High Level Design Validation and Test Workshop, 2009. HLDVT 2009. IEEE International},
  pages={166--171},
  year={2009},
  organization={IEEE}
}

@inproceedings{parveen2017hybrid,
  title={Hybrid Polymorphic Logic Gate with 5-Terminal Magnetic Domain Wall Motion Device},
  author={Parveen, Farhana and He, Zhezhi and Angizi, Shaahin and Fan, Deliang},
  booktitle={VLSI (ISVLSI), 2017 IEEE Computer Society Annual Symposium on},
  pages={152--157},
  year={2017},
  organization={IEEE}
}

@article{lecun1995convolutional,
  title={Convolutional networks for images, speech, and time series},
  author={LeCun, Yann and Bengio, Yoshua and others},
  journal={The handbook of brain theory and neural networks},
  volume={3361},
  number={10},
  pages={1995},
  year={1995}
}

@article{krizhevsky2010cifar,
  title={Cifar-10 (canadian institute for advanced research)},
  author={Krizhevsky, Alex and Nair, Vinod and Hinton, Geoffrey},
  journal={URL http://www. cs. toronto. edu/kriz/cifar. html},
  year={2010}
}

@inproceedings{netzer2011reading,
  title={Reading digits in natural images with unsupervised feature learning},
  author={Netzer, Yuval and Wang, Tao and Coates, Adam and Bissacco, Alessandro and Wu, Bo and Ng, Andrew Y},
  booktitle={NIPS workshop on deep learning and unsupervised feature learning},
  volume={2011},
  number={2},
  pages={5},
  year={2011}
}

@article{lecun2015lenet,
  title={LeNet-5, convolutional neural networks},
  author={LeCun, Yann and others},
  journal={URL: http://yann. lecun. com/exdb/lenet},
  pages={20},
  year={2015}
}

@article{xu2017can,
  title={Can you fool AI with adversarial examples on a visual Turing test?},
  author={Xu, Xiaojun and Chen, Xinyun and Liu, Chang and Rohrbach, Anna and Darell, Trevor and Song, Dawn},
  journal={arXiv preprint arXiv:1709.08693},
  year={2017}
}

@article{metzen2017universal,
  title={Universal adversarial perturbations against semantic image segmentation},
  author={Metzen, Jan Hendrik and Kumar, Mummadi Chaithanya and Brox, Thomas and Fischer, Volker},
  journal={stat},
  volume={1050},
  pages={19},
  year={2017}
}

@inproceedings{carlini2017towards,
  title={Towards evaluating the robustness of neural networks},
  author={Carlini, Nicholas and Wagner, David},
  booktitle={Security and Privacy (SP), 2017 IEEE Symposium on},
  pages={39--57},
  year={2017},
  organization={IEEE}
}

@article{Chen2018EADEA,
  title={EAD: Elastic-Net Attacks to Deep Neural Networks via Adversarial Examples},
  author={Pin-Yu Chen and Yash Sharma and Huan Zhang and Jinfeng Yi and Cho-Jui Hsieh},
  journal={CoRR},
  year={2018},
  volume={abs/1709.04114}
}



@article{zhou2016dorefa,
  title={DoReFa-Net: Training low bitwidth convolutional neural networks with low bitwidth gradients},
  author={Zhou, Shuchang and Wu, Yuxin and Ni, Zekun and Zhou, Xinyu and Wen, He and Zou, Yuheng},
  journal={arXiv preprint arXiv:1606.06160},
  year={2016}
}



@incollection{NIPS2017_6749,
title = {TernGrad: Ternary Gradients to Reduce Communication in Distributed Deep Learning},
author = {Wen, Wei and Xu, Cong and Yan, Feng and Wu, Chunpeng and Wang, Yandan and Chen, Yiran and Li, Hai},
booktitle = {Advances in Neural Information Processing Systems 30},
editor = {I. Guyon and U. V. Luxburg and S. Bengio and H. Wallach and R. Fergus and S. Vishwanathan and R. Garnett},
pages = {1509--1519},
year = {2017},
publisher = {Curran Associates, Inc.},
url = {http://papers.nips.cc/paper/6749-terngrad-ternary-gradients-to-reduce-communication-in-distributed-deep-learning.pdf}
}

@inproceedings{chen2015deepdriving,
  title={Deepdriving: Learning affordance for direct perception in autonomous driving},
  author={Chen, Chenyi and Seff, Ari and Kornhauser, Alain and Xiao, Jianxiong},
  booktitle={Computer Vision (ICCV), 2015 IEEE International Conference on},
  pages={2722--2730},
  year={2015},
  organization={IEEE}
}

@article{bahdanau2014neural,
  title={Neural machine translation by jointly learning to align and translate},
  author={Bahdanau, Dzmitry and Cho, Kyunghyun and Bengio, Yoshua},
  journal={arXiv preprint arXiv:1409.0473},
  year={2014}
}

@article{courbariaux2016binarized,
  title={Binarized neural networks: Training deep neural networks with weights and activations constrained to+ 1 or-1},
  author={Courbariaux, Matthieu and Hubara, Itay and Soudry, Daniel and El-Yaniv, Ran and Bengio, Yoshua},
  journal={arXiv preprint arXiv:1602.02830},
  year={2016}
}



@article{hinton2012neural,
  title={Neural networks for machine learning},
  author={Hinton, Geoffrey and Srivastava, Nitsh and Swersky, Kevin},
  journal={Coursera, video lectures},
  volume={264},
  year={2012}
}

@article{li2016ternary,
  title={Ternary weight networks},
  author={Li, Fengfu and Zhang, Bo and Liu, Bin},
  journal={arXiv preprint arXiv:1605.04711},
  year={2016}
}
@inproceedings{courbariaux2015binaryconnect,
  title={Binaryconnect: Training deep neural networks with binary weights during propagations},
  author={Courbariaux, Matthieu and Bengio, Yoshua and David, Jean-Pierre},
  booktitle={Advances in neural information processing systems},
  pages={3123--3131},
  year={2015}
}


@article{han2015deep,
  title={Deep compression: Compressing deep neural networks with pruning, trained quantization and huffman coding},
  author={Han, Song and Mao, Huizi and Dally, William J},
  journal={arXiv preprint arXiv:1510.00149},
  year={2015}
}

@inproceedings{rastegari2016xnor,
  title={Xnor-net: Imagenet classification using binary convolutional neural networks},
  author={Rastegari, Mohammad and Ordonez, Vicente and Redmon, Joseph and Farhadi, Ali},
  booktitle={European Conference on Computer Vision},
  pages={525--542},
  year={2016},
  organization={Springer}
}

@article{wald1945statistical,
  title={Statistical decision functions which minimize the maximum risk},
  author={Wald, Abraham},
  journal={Annals of Mathematics},
  pages={265--280},
  year={1945},
  publisher={JSTOR}
}


@article{Santhanam2018DefendingAA,
  title={Defending Against Adversarial Attacks by Leveraging an Entire GAN},
  author={Gokula Krishnan Santhanam and Paulina Grnarova},
  journal={CoRR},
  year={2018},
  volume={abs/1805.10652}
}

@inproceedings{
samangouei2018defensegan,
title={Defense-{GAN}: Protecting Classifiers Against Adversarial Attacks Using Generative Models},
author={Pouya Samangouei and Maya Kabkab and Rama Chellappa},
booktitle={International Conference on Learning Representations},
year={2018},
url={https://openreview.net/forum?id=BkJ3ibb0-},
}
@inproceedings{
raghunathan2018certified,
title={Certified Defenses against Adversarial Examples },
author={Aditi Raghunathan and Jacob Steinhardt and Percy Liang},
booktitle={International Conference on Learning Representations},
year={2018},
url={https://openreview.net/forum?id=Bys4ob-Rb},
}

@inproceedings{
s.2018stochastic,
title={Stochastic activation pruning for robust adversarial defense},
author={Guneet S. Dhillon and Kamyar Azizzadenesheli and Jeremy D. Bernstein and Jean Kossaifi and Aran Khanna and Zachary C. Lipton and Animashree Anandkumar},
booktitle={International Conference on Learning Representations},
year={2018},
url={https://openreview.net/forum?id=H1uR4GZRZ},
}
@inproceedings{prakash2018deflecting,
  title={Deflecting adversarial attacks with pixel deflection},
  author={Prakash, Aaditya and Moran, Nick and Garber, Solomon and DiLillo, Antonella and Storer, James},
  booktitle={Proceedings of the IEEE Conference on Computer Vision and Pattern Recognition},
  pages={8571--8580},
  year={2018}
}
@article{maaten2008visualizing,
  title={Visualizing data using t-SNE},
  author={Maaten, Laurens van der and Hinton, Geoffrey},
  journal={Journal of machine learning research},
  volume={9},
  number={Nov},
  pages={2579--2605},
  year={2008}
}

@inproceedings{rozsa2018towards,
  title={Towards Robust Deep Neural Networks with BANG},
  author={Rozsa, Andras and Gunther, Manuel and Boult, Terrance E},
  booktitle={2018 IEEE Winter Conference on Applications of Computer Vision (WACV)},
  pages={803--811},
  year={2018},
  organization={IEEE}
}
@article{xu2015empirical,
  title={Empirical evaluation of rectified activations in convolutional network},
  author={Xu, Bing and Wang, Naiyan and Chen, Tianqi and Li, Mu},
  journal={arXiv preprint arXiv:1505.00853},
  year={2015}
}

@article{akhtar2018threat,
  title={Threat of Adversarial Attacks on Deep Learning in Computer Vision: A Survey},
  author={Akhtar, Naveed and Mian, Ajmal},
  journal={IEEE Access},
  volume={6},
  pages={14410--14430},
  year={2018},
  publisher={IEEE}
}



@InProceedings{pmlr-v80-uesato18a,
  title = 	 {Adversarial Risk and the Dangers of Evaluating Against Weak Attacks},
  author = 	 {Uesato, Jonathan and O'Donoghue, Brendan and Kohli, Pushmeet and van den Oord, Aaron},
  booktitle = 	 {Proceedings of the 35th International Conference on Machine Learning},
  pages = 	 {5025--5034},
  year = 	 {2018},
  editor = 	 {Dy, Jennifer and Krause, Andreas},
  volume = 	 {80},
  series = 	 {Proceedings of Machine Learning Research},
  address = 	 {Stockholmsmässan, Stockholm Sweden},
  month = 	 {10--15 Jul},
  publisher = 	 {PMLR},
  pdf = 	 {http://proceedings.mlr.press/v80/uesato18a/uesato18a.pdf},
  url = 	 {http://proceedings.mlr.press/v80/uesato18a.html},
  abstract = 	 {This paper investigates recently proposed approaches for defending against adversarial examples and evaluating adversarial robustness. We motivate \emph{adversarial risk} as an objective for achieving models robust to worst-case inputs. We then frame commonly used attacks and evaluation metrics as defining a tractable surrogate objective to the true adversarial risk. This suggests that models may optimize this surrogate rather than the true adversarial risk. We formalize this notion as \emph{obscurity to an adversary}, and develop tools and heuristics for identifying obscured models and designing transparent models. We demonstrate that this is a significant problem in practice by repurposing gradient-free optimization techniques into adversarial attacks, which we use to decrease the accuracy of several recently proposed defenses to near zero. Our hope is that our formulations and results will help researchers to develop more powerful defenses.}
}

@inproceedings{liao2018defense,
  title={Defense against adversarial attacks using high-level representation guided denoiser},
  author={Liao, Fangzhou and Liang, Ming and Dong, Yinpeng and Pang, Tianyu and Zhu, Jun and Hu, Xiaolin},
  booktitle={Proceedings of the IEEE Conference on Computer Vision and Pattern Recognition},
  pages={1778--1787},
  year={2018}
}

@article{athalye2018robustness,
  title={On the Robustness of the CVPR 2018 White-Box Adversarial Example Defenses},
  author={Athalye, Anish and Carlini, Nicholas},
  journal={arXiv preprint arXiv:1804.03286},
  year={2018}
}

@article{bengio2013estimating,
  title={Estimating or propagating gradients through stochastic neurons for conditional computation},
  author={Bengio, Yoshua and L{\'e}onard, Nicholas and Courville, Aaron},
  journal={arXiv preprint arXiv:1308.3432},
  year={2013}
}

@inproceedings{sun2018feature,
  title={Feature Quantization for Defending Against Distortion of Images},
  author={Sun, Zhun and Ozay, Mete and Zhang, Yan and Liu, Xing and Okatani, Takayuki},
  booktitle={Proceedings of the IEEE Conference on Computer Vision and Pattern Recognition},
  pages={7957--7966},
  year={2018}
}

@inproceedings{liu2018understanding,
  title={Understanding adversarial attack and defense towards deep compressed neural networks},
  author={Liu, Qi and Liu, Tao and Wen, Wujie},
  booktitle={Cyber Sensing 2018},
  volume={10630},
  pages={106300Q},
  year={2018},
  organization={International Society for Optics and Photonics}
}

@article{yoshida2017spectral,
  title={Spectral Norm Regularization for Improving the Generalizability of Deep Learning},
  author={Yoshida, Yuichi and Miyato, Takeru},
  journal={arXiv preprint arXiv:1705.10941},
  year={2017}
}

@inproceedings{    
anonymous2019random,    
title={RANDOM MASK: Towards Robust Convolutional Neural Networks},    
author={Anonymous},    
booktitle={Submitted to International Conference on Learning Representations},    
year={2019},    
url={https://openreview.net/forum?id=SkgkJn05YX},    
note={under review}    
}
@inproceedings{    
anonymous2019label,    
title={Label Smoothing and Logit Squeezing: A Replacement for Adversarial Training?},    
author={Anonymous},    
booktitle={Submitted to International Conference on Learning Representations},    
year={2019},    
url={https://openreview.net/forum?id=BJlr0j0ctX},    
note={under review}    
}
@inproceedings{ng2004feature,
  title={Feature selection, L 1 vs. L 2 regularization, and rotational invariance},
  author={Ng, Andrew Y},
  booktitle={Proceedings of the twenty-first international conference on Machine learning},
  pages={78},
  year={2004},
  organization={ACM}
}

@article{JMLR:v15:srivastava14a,
  author  = {Nitish Srivastava and Geoffrey Hinton and Alex Krizhevsky and Ilya Sutskever and Ruslan Salakhutdinov},
  title   = {Dropout: A Simple Way to Prevent Neural Networks from Overfitting},
  journal = {Journal of Machine Learning Research},
  year    = {2014},
  volume  = {15},
  pages   = {1929-1958},
  url     = {http://jmlr.org/papers/v15/srivastava14a.html}
}

@article{xiao2017fashion,
  title={Fashion-mnist: a novel image dataset for benchmarking machine learning algorithms},
  author={Xiao, Han and Rasul, Kashif and Vollgraf, Roland},
  journal={arXiv preprint arXiv:1708.07747},
  year={2017}
}

@techreport{krizhevsky2009learning,
  title={Learning multiple layers of features from tiny images},
  author={Krizhevsky, Alex and Hinton, Geoffrey},
  year={2009},
  institution={Citeseer}
}


@article{lecun1998gradient,
  title={Gradient-based learning applied to document recognition},
  author={LeCun, Yann and Bottou, L{\'e}on and Bengio, Yoshua and Haffner, Patrick},
  journal={Proceedings of the IEEE},
  volume={86},
  number={11},
  pages={2278--2324},
  year={1998},
  publisher={IEEE}
}

@article{kingma2014adam,
  title={Adam: A method for stochastic optimization},
  author={Kingma, Diederik P and Ba, Jimmy},
  journal={arXiv preprint arXiv:1412.6980},
  year={2014}
}

@article{lecuyer2018certified,
  title={Certified Robustness to Adversarial Examples with Differential Privacy},
  author={Lecuyer, M and Atlidakis, V and Geambasu, R and Hsu, D and Jana, S},
  journal={ArXiv e-prints},
  year={2018}
}

@article{dhillon2018stochastic,
  title={Stochastic activation pruning for robust adversarial defense},
  author={Dhillon, Guneet S and Azizzadenesheli, Kamyar and Lipton, Zachary C and Bernstein, Jeremy and Kossaifi, Jean and Khanna, Aran and Anandkumar, Anima},
  journal={arXiv preprint arXiv:1803.01442},
  year={2018}
}


@inproceedings{wan2013regularization,
  title={Regularization of neural networks using dropconnect},
  author={Wan, Li and Zeiler, Matthew and Zhang, Sixin and Le Cun, Yann and Fergus, Rob},
  booktitle={International Conference on Machine Learning},
  pages={1058--1066},
  year={2013}
}

@inproceedings{    
anonymous2019l2-nonexpansive,    
title={L2-Nonexpansive Neural Networks},    
author={Anonymous},    
booktitle={Submitted to International Conference on Learning Representations},    
year={2019},    
url={https://openreview.net/forum?id=ByxGSsR9FQ},    
note={under review}    
}

@article{DBLP:journals/corr/abs-1804-03286,
  author    = {Anish Athalye and
               Nicholas Carlini},
  title     = {On the Robustness of the {CVPR} 2018 White-Box Adversarial Example
               Defenses},
  journal   = {CoRR},
  volume    = {abs/1804.03286},
  year      = {2018},
  url       = {http://arxiv.org/abs/1804.03286},
  archivePrefix = {arXiv},
  eprint    = {1804.03286},
  timestamp = {Mon, 13 Aug 2018 16:47:37 +0200},
  biburl    = {https://dblp.org/rec/bib/journals/corr/abs-1804-03286},
  bibsource = {dblp computer science bibliography, https://dblp.org}
}

@article{liu2018adv,
  title={Adv-BNN: Improved Adversarial Defense through Robust Bayesian Neural Network},
  author={Liu, Xuanqing and Li, Yao and Wu, Chongruo and Hsieh, Cho-Jui},
  journal={International Conference on Learning Representations (ICLR)},
  year={2019}
}

@misc{hinton_lecture,
  author = {geoffrey hinton},
  title = {Neural Networks for Machine Learning: Overview of ways to improve generalization},
  howpublished = "\url{https://www.cs.toronto.edu/~tijmen/csc321/slides/lecture_slides_lec9.pdf}",
  year = {2014}, 
}
@article{tibshirani1996regression,
  title={Regression shrinkage and selection via the lasso},
  author={Tibshirani, Robert},
  journal={Journal of the Royal Statistical Society: Series B (Methodological)},
  volume={58},
  number={1},
  pages={267--288},
  year={1996},
  publisher={Wiley Online Library}
}

@inproceedings{he2019PNI,
 title={Parametric Noise Injection: Trainable Randomness to Improve Deep Neural Network Robustness against Adversarial Attack},
 author={He, Zhezhi and Adnan Siraj Rakin and Fan, Deliang},
 booktitle={Proceedings of the IEEE Conference on Computer Vision and Pattern Recognition},
 pages={},
 year={2019}
}

@article{ye2019second,
  title={Second Rethinking of Network Pruning in the Adversarial Setting},
  author={Ye, Shaokai and Xu, Kaidi and Liu, Sijia and Cheng, Hao and Lambrechts, Jan-Henrik and Zhang, Huan and Zhou, Aojun and Ma, Kaisheng and Wang, Yanzhi and Lin, Xue},
  journal={arXiv preprint arXiv:1903.12561},
  year={2019}
}

@inproceedings{he2017channel,
  title={Channel pruning for accelerating very deep neural networks},
  author={He, Yihui and Zhang, Xiangyu and Sun, Jian},
  booktitle={Proceedings of the IEEE International Conference on Computer Vision},
  pages={1389--1397},
  year={2017}
}

@article{wang2018structured,
  title={Structured Pruning for Efficient ConvNets via Incremental Regularization},
  author={Wang, Huan and Zhang, Qiming and Wang, Yuehai and Hu, Haoji},
  journal={arXiv preprint arXiv:1811.08390},
  year={2018}
}

@misc{
bietti*2019on,
title={On Regularization and Robustness of Deep Neural Networks},
author={Alberto Bietti* and Grégoire Mialon* and Julien Mairal},
year={2019},
url={https://openreview.net/forum?id=HkMlGnC9KQ},
}
@inproceedings{
lin2018defensive,
title={Defensive Quantization: When Efficiency Meets Robustness},
author={Ji Lin and Chuang Gan and Song Han},
booktitle={International Conference on Learning Representations},
year={2019},
url={https://openreview.net/forum?id=ryetZ20ctX},
}



@inproceedings{guo2018sparse,
  title={Sparse dnns with improved adversarial robustness},
  author={Guo, Yiwen and Zhang, Chao and Zhang, Changshui and Chen, Yurong},
  booktitle={Advances in neural information processing systems},
  pages={242--251},
  year={2018}
}

@misc{
  ye2018defending,
  title={Defending DNN Adversarial Attacks with Pruning and Logits Augmentation},
  author={Shaokai Ye and Siyue Wang and Xiao Wang and Bo Yuan and Wujie Wen and Xue Lin},
  year={2018},
  url={https://openreview.net/forum?id=S1qI2FJDM}
}

@inproceedings{he2019simultaneously,
 title={Parametric Noise Injection: Trainable Randomness to Improve Deep Neural Network Robustness against Adversarial Attack},
 author={He, Zhezhi and Fan, Deliang},
 booktitle={Proceedings of the IEEE Conference on Computer Vision and Pattern Recognition},
 pages={},
 year={2019}
}

@article{rakin2018defend,
  title={Defend deep neural networks against adversarial examples via fixed anddynamic quantized activation functions},
  author={Rakin, Adnan Siraj and Yi, Jinfeng and Gong, Boqing and Fan, Deliang},
  journal={arXiv preprint arXiv:1807.06714},
  year={2018}
}

@inproceedings{wen2016learning,
  title={Learning structured sparsity in deep neural networks},
  author={Wen, Wei and Wu, Chunpeng and Wang, Yandan and Chen, Yiran and Li, Hai},
  booktitle={Advances in neural information processing systems},
  pages={2074--2082},
  year={2016}
}

\begin{thebibliography}{39}
\providecommand{\natexlab}[1]{#1}
\providecommand{\url}[1]{\texttt{#1}}
\expandafter\ifx\csname urlstyle\endcsname\relax
  \providecommand{\doi}[1]{doi: #1}\else
  \providecommand{\doi}{doi: \begingroup \urlstyle{rm}\Url}\fi

\bibitem[Akhtar and Mian(2018)]{akhtar2018threat}
N.~Akhtar and A.~Mian.
\newblock Threat of adversarial attacks on deep learning in computer vision: A
  survey.
\newblock \emph{IEEE Access}, 6:\penalty0 14410--14430, 2018.

\bibitem[Athalye et~al.(2018)Athalye, Carlini, and Wagner]{pmlr-v80-athalye18a}
A.~Athalye, N.~Carlini, and D.~Wagner.
\newblock Obfuscated gradients give a false sense of security: Circumventing
  defenses to adversarial examples.
\newblock In J.~Dy and A.~Krause, editors, \emph{Proceedings of the 35th
  International Conference on Machine Learning}, volume~80 of \emph{Proceedings
  of Machine Learning Research}, pages 274--283, Stockholmsmässan, Stockholm
  Sweden, 10--15 Jul 2018. PMLR.
\newblock URL \url{http://proceedings.mlr.press/v80/athalye18a.html}.

\bibitem[Chen et~al.(2017{\natexlab{a}})Chen, Zhang, Chen, Yi, and
  Hsieh]{chen2017show}
H.~Chen, H.~Zhang, P.-Y. Chen, J.~Yi, and C.-J. Hsieh.
\newblock Show-and-fool: Crafting adversarial examples for neural image
  captioning.
\newblock \emph{arXiv preprint arXiv:1712.02051}, 2017{\natexlab{a}}.

\bibitem[Chen et~al.(2017{\natexlab{b}})Chen, Zhang, Sharma, Yi, and
  Hsieh]{chen2017zoo}
P.-Y. Chen, H.~Zhang, Y.~Sharma, J.~Yi, and C.-J. Hsieh.
\newblock Zoo: Zeroth order optimization based black-box attacks to deep neural
  networks without training substitute models.
\newblock In \emph{Proceedings of the 10th ACM Workshop on Artificial
  Intelligence and Security}, pages 15--26. ACM, 2017{\natexlab{b}}.

\bibitem[Courbariaux et~al.(2015)Courbariaux, Bengio, and
  David]{courbariaux2015binaryconnect}
M.~Courbariaux, Y.~Bengio, and J.-P. David.
\newblock Binaryconnect: Training deep neural networks with binary weights
  during propagations.
\newblock In \emph{Advances in neural information processing systems}, pages
  3123--3131, 2015.

\bibitem[Courbariaux et~al.(2016)Courbariaux, Hubara, Soudry, El-Yaniv, and
  Bengio]{courbariaux2016binarized}
M.~Courbariaux, I.~Hubara, D.~Soudry, R.~El-Yaniv, and Y.~Bengio.
\newblock Binarized neural networks: Training deep neural networks with weights
  and activations constrained to+ 1 or-1.
\newblock \emph{arXiv preprint arXiv:1602.02830}, 2016.

\bibitem[Dhillon et~al.(2018)Dhillon, Azizzadenesheli, Bernstein, Kossaifi,
  Khanna, Lipton, and Anandkumar]{s.2018stochastic}
G.~S. Dhillon, K.~Azizzadenesheli, J.~D. Bernstein, J.~Kossaifi, A.~Khanna,
  Z.~C. Lipton, and A.~Anandkumar.
\newblock Stochastic activation pruning for robust adversarial defense.
\newblock In \emph{International Conference on Learning Representations}, 2018.
\newblock URL \url{https://openreview.net/forum?id=H1uR4GZRZ}.

\bibitem[Goodfellow et~al.(2014)Goodfellow, Shlens, and
  Szegedy]{goodfellow2014explaining}
I.~J. Goodfellow, J.~Shlens, and C.~Szegedy.
\newblock Explaining and harnessing adversarial examples.
\newblock \emph{arXiv preprint arXiv:1412.6572}, 2014.

\bibitem[Guo et~al.(2018)Guo, Zhang, Zhang, and Chen]{guo2018sparse}
Y.~Guo, C.~Zhang, C.~Zhang, and Y.~Chen.
\newblock Sparse dnns with improved adversarial robustness.
\newblock In \emph{Advances in neural information processing systems}, pages
  242--251, 2018.

\bibitem[Han et~al.(2015{\natexlab{a}})Han, Mao, and Dally]{han2015deep}
S.~Han, H.~Mao, and W.~J. Dally.
\newblock Deep compression: Compressing deep neural networks with pruning,
  trained quantization and huffman coding.
\newblock \emph{arXiv preprint arXiv:1510.00149}, 2015{\natexlab{a}}.

\bibitem[Han et~al.(2015{\natexlab{b}})Han, Pool, Tran, and
  Dally]{han2015learning}
S.~Han, J.~Pool, J.~Tran, and W.~Dally.
\newblock Learning both weights and connections for efficient neural network.
\newblock In \emph{Advances in neural information processing systems}, pages
  1135--1143, 2015{\natexlab{b}}.

\bibitem[Han et~al.(2016)Han, Shen, Philipose, Agarwal, Wolman, and
  Krishnamurthy]{han2016mcdnn}
S.~Han, H.~Shen, M.~Philipose, S.~Agarwal, A.~Wolman, and A.~Krishnamurthy.
\newblock Mcdnn: An approximation-based execution framework for deep stream
  processing under resource constraints.
\newblock In \emph{Proceedings of the 14th Annual International Conference on
  Mobile Systems, Applications, and Services}, pages 123--136. ACM, 2016.

\bibitem[He et~al.(2016)He, Zhang, Ren, and Sun]{he2016deep}
K.~He, X.~Zhang, S.~Ren, and J.~Sun.
\newblock Deep residual learning for image recognition.
\newblock In \emph{Proceedings of the IEEE conference on computer vision and
  pattern recognition}, pages 770--778, 2016.

\bibitem[He et~al.(2017)He, Zhang, and Sun]{he2017channel}
Y.~He, X.~Zhang, and J.~Sun.
\newblock Channel pruning for accelerating very deep neural networks.
\newblock In \emph{Proceedings of the IEEE International Conference on Computer
  Vision}, pages 1389--1397, 2017.

\bibitem[He and Fan(2019)]{he2019simultaneously}
Z.~He and D.~Fan.
\newblock Parametric noise injection: Trainable randomness to improve deep
  neural network robustness against adversarial attack.
\newblock In \emph{Proceedings of the IEEE Conference on Computer Vision and
  Pattern Recognition}, 2019.

\bibitem[He et~al.(2019)He, Rakin, and Fan]{he2019PNI}
Z.~He, A.~S. Rakin, and D.~Fan.
\newblock Parametric noise injection: Trainable randomness to improve deep
  neural network robustness against adversarial attack.
\newblock In \emph{Proceedings of the IEEE Conference on Computer Vision and
  Pattern Recognition}, 2019.

\bibitem[Hinton et~al.(2012{\natexlab{a}})Hinton, Deng, Yu, Dahl, Mohamed,
  Jaitly, Senior, Vanhoucke, Nguyen, Sainath, et~al.]{hinton2012deep}
G.~Hinton, L.~Deng, D.~Yu, G.~E. Dahl, A.-r. Mohamed, N.~Jaitly, A.~Senior,
  V.~Vanhoucke, P.~Nguyen, T.~N. Sainath, et~al.
\newblock Deep neural networks for acoustic modeling in speech recognition: The
  shared views of four research groups.
\newblock \emph{IEEE Signal Processing Magazine}, 29\penalty0 (6):\penalty0
  82--97, 2012{\natexlab{a}}.

\bibitem[Hinton et~al.(2012{\natexlab{b}})Hinton, Srivastava, and
  Swersky]{hinton2012neural}
G.~Hinton, N.~Srivastava, and K.~Swersky.
\newblock Neural networks for machine learning.
\newblock \emph{Coursera, video lectures}, 264, 2012{\natexlab{b}}.

\bibitem[Hung et~al.(2017)Hung, Chen, Lai, Lin, and Lee]{hung2017comparing}
C.-Y. Hung, W.-C. Chen, P.-T. Lai, C.-H. Lin, and C.-C. Lee.
\newblock Comparing deep neural network and other machine learning algorithms
  for stroke prediction in a large-scale population-based electronic medical
  claims database.
\newblock In \emph{2017 39th Annual International Conference of the IEEE
  Engineering in Medicine and Biology Society (EMBC)}, pages 3110--3113. IEEE,
  2017.

\bibitem[Lecuyer et~al.(2018{\natexlab{a}})Lecuyer, Atlidakis, Geambasu, Hsu,
  and Jana]{lecuyer2018certified}
M.~Lecuyer, V.~Atlidakis, R.~Geambasu, D.~Hsu, and S.~Jana.
\newblock Certified robustness to adversarial examples with differential
  privacy.
\newblock \emph{ArXiv e-prints}, 2018{\natexlab{a}}.

\bibitem[Lecuyer et~al.(2018{\natexlab{b}})Lecuyer, Atlidakis, Geambasu, Hsu,
  and Jana]{lecuyer2018connection}
M.~Lecuyer, V.~Atlidakis, R.~Geambasu, D.~Hsu, and S.~Jana.
\newblock On the connection between differential privacy and adversarial
  robustness in machine learning.
\newblock \emph{arXiv preprint arXiv:1802.03471}, 2018{\natexlab{b}}.

\bibitem[Lin et~al.(2019)Lin, Gan, and Han]{lin2018defensive}
J.~Lin, C.~Gan, and S.~Han.
\newblock Defensive quantization: When efficiency meets robustness.
\newblock In \emph{International Conference on Learning Representations}, 2019.
\newblock URL \url{https://openreview.net/forum?id=ryetZ20ctX}.

\bibitem[Liu et~al.(2017)Liu, Cheng, Zhang, and Hsieh]{liu2017towards}
X.~Liu, M.~Cheng, H.~Zhang, and C.-J. Hsieh.
\newblock Towards robust neural networks via random self-ensemble.
\newblock \emph{arXiv preprint arXiv:1712.00673}, 2017.

\bibitem[Liu et~al.(2016)Liu, Chen, Liu, and Song]{liu2016delving}
Y.~Liu, X.~Chen, C.~Liu, and D.~Song.
\newblock Delving into transferable adversarial examples and black-box attacks.
\newblock \emph{arXiv preprint arXiv:1611.02770}, 2016.

\bibitem[Madry et~al.(2018)Madry, Makelov, Schmidt, Tsipras, and
  Vladu]{madry2018towards}
A.~Madry, A.~Makelov, L.~Schmidt, D.~Tsipras, and A.~Vladu.
\newblock Towards deep learning models resistant to adversarial attacks.
\newblock In \emph{International Conference on Learning Representations}, 2018.
\newblock URL \url{https://openreview.net/forum?id=rJzIBfZAb}.

\bibitem[Molchanov et~al.(2017)Molchanov, Ashukha, and
  Vetrov]{molchanov2017variational}
D.~Molchanov, A.~Ashukha, and D.~Vetrov.
\newblock Variational dropout sparsifies deep neural networks.
\newblock In \emph{Proceedings of the 34th International Conference on Machine
  Learning-Volume 70}, pages 2498--2507. JMLR. org, 2017.

\bibitem[Raghunathan et~al.(2018)Raghunathan, Steinhardt, and
  Liang]{raghunathan2018certified}
A.~Raghunathan, J.~Steinhardt, and P.~Liang.
\newblock Certified defenses against adversarial examples.
\newblock In \emph{International Conference on Learning Representations}, 2018.
\newblock URL \url{https://openreview.net/forum?id=Bys4ob-Rb}.

\bibitem[Rakin et~al.(2018)Rakin, Yi, Gong, and Fan]{rakin2018defend}
A.~S. Rakin, J.~Yi, B.~Gong, and D.~Fan.
\newblock Defend deep neural networks against adversarial examples via fixed
  anddynamic quantized activation functions.
\newblock \emph{arXiv preprint arXiv:1807.06714}, 2018.

\bibitem[Samangouei et~al.(2018)Samangouei, Kabkab, and
  Chellappa]{samangouei2018defensegan}
P.~Samangouei, M.~Kabkab, and R.~Chellappa.
\newblock Defense-{GAN}: Protecting classifiers against adversarial attacks
  using generative models.
\newblock In \emph{International Conference on Learning Representations}, 2018.
\newblock URL \url{https://openreview.net/forum?id=BkJ3ibb0-}.

\bibitem[Simonyan and Zisserman(2014)]{simonyan2014very}
K.~Simonyan and A.~Zisserman.
\newblock Very deep convolutional networks for large-scale image recognition.
\newblock \emph{arXiv preprint arXiv:1409.1556}, 2014.

\bibitem[Szegedy et~al.(2013)Szegedy, Zaremba, Sutskever, Bruna, Erhan,
  Goodfellow, and Fergus]{Szegedy2013IntriguingPO}
C.~Szegedy, W.~Zaremba, I.~Sutskever, J.~Bruna, D.~Erhan, I.~J. Goodfellow, and
  R.~Fergus.
\newblock Intriguing properties of neural networks.
\newblock \emph{CoRR}, abs/1312.6199, 2013.

\bibitem[Tibshirani(1996)]{tibshirani1996regression}
R.~Tibshirani.
\newblock Regression shrinkage and selection via the lasso.
\newblock \emph{Journal of the Royal Statistical Society: Series B
  (Methodological)}, 58\penalty0 (1):\penalty0 267--288, 1996.

\bibitem[Wang et~al.(2018)Wang, Zhang, Wang, and Hu]{wang2018structured}
H.~Wang, Q.~Zhang, Y.~Wang, and H.~Hu.
\newblock Structured pruning for efficient convnets via incremental
  regularization.
\newblock \emph{arXiv preprint arXiv:1811.08390}, 2018.

\bibitem[Wen et~al.(2016)Wen, Wu, Wang, Chen, and Li]{wen2016learning}
W.~Wen, C.~Wu, Y.~Wang, Y.~Chen, and H.~Li.
\newblock Learning structured sparsity in deep neural networks.
\newblock In \emph{Advances in neural information processing systems}, pages
  2074--2082, 2016.

\bibitem[Xie et~al.(2018)Xie, Wang, Zhang, Ren, and Yuille]{xie2018mitigating}
C.~Xie, J.~Wang, Z.~Zhang, Z.~Ren, and A.~Yuille.
\newblock Mitigating adversarial effects through randomization.
\newblock In \emph{International Conference on Learning Representations}, 2018.
\newblock URL \url{https://openreview.net/forum?id=Sk9yuql0Z}.

\bibitem[Ye et~al.(2018)Ye, Wang, Wang, Yuan, Wen, and Lin]{ye2018defending}
S.~Ye, S.~Wang, X.~Wang, B.~Yuan, W.~Wen, and X.~Lin.
\newblock Defending dnn adversarial attacks with pruning and logits
  augmentation, 2018.
\newblock URL \url{https://openreview.net/forum?id=S1qI2FJDM}.

\bibitem[Ye et~al.(2019)Ye, Xu, Liu, Cheng, Lambrechts, Zhang, Zhou, Ma, Wang,
  and Lin]{ye2019second}
S.~Ye, K.~Xu, S.~Liu, H.~Cheng, J.-H. Lambrechts, H.~Zhang, A.~Zhou, K.~Ma,
  Y.~Wang, and X.~Lin.
\newblock Second rethinking of network pruning in the adversarial setting.
\newblock \emph{arXiv preprint arXiv:1903.12561}, 2019.

\bibitem[Yoshida and Miyato(2017)]{yoshida2017spectral}
Y.~Yoshida and T.~Miyato.
\newblock Spectral norm regularization for improving the generalizability of
  deep learning.
\newblock \emph{arXiv preprint arXiv:1705.10941}, 2017.

\bibitem[Zhou et~al.(2016)Zhou, Wu, Ni, Zhou, Wen, and Zou]{zhou2016dorefa}
S.~Zhou, Y.~Wu, Z.~Ni, X.~Zhou, H.~Wen, and Y.~Zou.
\newblock Dorefa-net: Training low bitwidth convolutional neural networks with
  low bitwidth gradients.
\newblock \emph{arXiv preprint arXiv:1606.06160}, 2016.

\end{thebibliography}
}

\end{document}